\documentclass[10pt,twocolumn,letterpaper]{article}

\usepackage[pagenumbers]{cvpr} 

\usepackage{graphicx}
\usepackage{amsmath}
\usepackage{amssymb}
\usepackage{booktabs}
\usepackage{enumitem}
\makeatletter
\@namedef{ver@everyshi.sty}{}
\makeatother
\usepackage{tikz}
\usepackage{tikz,pgfplots}
    \usepgfplotslibrary{groupplots}
    \usetikzlibrary{arrows.meta}
    \usetikzlibrary{tikzmark}
\DeclareMathOperator*{\argmin}{arg\,min}

\usepackage[pagebackref,breaklinks,colorlinks]{hyperref}

\usepackage[capitalize]{cleveref}
\crefname{section}{Sec.}{Secs.}
\Crefname{section}{Section}{Sections}
\Crefname{table}{Table}{Tables}
\crefname{table}{Tab.}{Tabs.}

\def\cvprPaperID{9778} 
\def\confName{CVPR}
\def\confYear{2023}

\newcommand\decoder{\mathrm{D}}
\newcommand\weights{\mathbf{W}}
\newcommand\inimg{Z}
\newcommand\gaussiankernel{K}
\newcommand\numchannels{c}
\newcommand\width{w}
\newcommand\height{h}
\newcommand\ogposenc{\Phi}
\newcommand\posenc{\phi}
\newcommand\posfreq{l}
\newcommand\loss{\mathcal{L}}
\newcommand\reconloss{\mathcal{L_\text{recon}}}
\newcommand\spatialloss{\mathcal{L_\text{spatial}}}
\newcommand\lossweight{\alpha}
\newcommand\img{I}
\newcommand\targetimage{I_\text{target}}
\newcommand\lightness{I_\text{L}}
\newcommand\offset{\gamma}
\newcommand{\norm}[1]{\left\lVert#1\right\rVert}
\newcommand\real{\mathbb{R}}
\newcommand\chmaps{C}
\newcommand\pixelfeat{p}
\newcommand\ddnum{m}
\newcommand\targetchan{n}
\newcommand\integers{\mathbb{Z}}
\newcommand\blurfactor{b}
\newcommand\blurstd{\sigma}

\begin{document}

\title{Unsupervised Superpixel Generation using Edge-Sparse Embedding}

\author{Jakob Geusen\textsuperscript{1}, Gustav Bredell\textsuperscript{2}, Tianfei Zhou\textsuperscript{2}, Ender Konukoglu\textsuperscript{2}\\
Department of Information Technology and Electrical Engineering\\
ETH-Zurich, Zurich, Switzerland\\
\textsuperscript{1}{\tt\small jgeusen@student.ethz.ch}\\
\textsuperscript{2}{\tt\small \{gustav.bredell, tianfei.zhou, ender.konukoglu\}@vision.ee.ethz.ch}\\
}
\maketitle

\begin{abstract}
    Partitioning an image into superpixels based on the similarity of pixels with respect to features such as colour or spatial location can significantly reduce data complexity and improve subsequent image processing tasks. Initial algorithms for unsupervised superpixel generation solely relied on local cues without prioritizing significant edges over arbitrary ones.
    On the other hand, more recent methods based on unsupervised deep learning either fail to properly address the trade-off between superpixel edge adherence and compactness or lack control over the generated number of superpixels. 
    By using random images with strong spatial correlation as input, \ie, blurred noise images, in a non-convolutional image decoder we can reduce the expected number of contrasts and enforce smooth, connected edges in the reconstructed image.
    We generate edge-sparse pixel embeddings by encoding additional spatial information into the piece-wise smooth activation maps from the decoder's last hidden layer and use a standard clustering algorithm to extract high quality superpixels. Our proposed method reaches state-of-the-art performance on the BSDS500, PASCAL-Context and a microscopy dataset.
\end{abstract}

\section{Introduction}
\label{sec:intro}

Superpixel algorithms split an image into patches based on the similarity of pixels with respect to certain features such as colour or position. The subsequent representation can be used as pre-processing step for various tasks such as object detection \cite{yan_object_2015, sultani_automatic_2018}, hyperspectral image processing \cite{subudhi_survey_2021} and saliency detection \cite{liu_superpixel-based_2014, liu_saliency_2017, fang_novel_2018}. Furthermore, the grouping of pixels into superpixels can also be seen as a first step towards obtaining a segmentation, where the final segmentation is obtained by superpixel merging \cite{lei_superpixel-based_2019, gadde_superpixel_2016}. The usefulness of superpixel generation as a pre-processing step depends on several factors such as the compactness of the pixels or how well it aligns with object boundaries in the image.

The most prominent types of classical approaches for superpixel generation are graph and clustering based methods. The graph based methods interpret an image's pixels as nodes and assign each edge between neighbouring pixels a similarity measure \cite{felzenszwalb_efficient_2004, liu_entropy_2011}. Strongly connected subgraphs can then be extracted and transformed to superpixels. A very naive and inefficient idea to approach superpixel generation is by applying k-means clustering on the RGB values. SLIC \cite{achanta_slic_2012} addresses the inefficiency of applying k-means globally by limiting the pixel cluster assignment to pixels in a cluster's neighbourhood. Moreover, it transforms the RGB values into the $lab$ color space and concatenates $xy$ coordinates to the pixel vectors to be clustered. Since then multiple improvements have been proposed. SNIC \cite{achanta_superpixels_2017} is a non-iterative version of SLIC enabling faster execution while requiring less memory. Power-SLIC \cite{fiedler_power-slic_2021} incorporates generalized power diagrams to produce regular shaped superpixels that are noise robust. As opposed to the SLIC based methods linear spectral clustering (LSC) \cite{zhengqin_li_superpixel_2015} uses a kernel to cluster in a higher dimensional space to ensure correct superpixel extraction from images with high intensity variability. Furthermore, the use of a local fuzzy clustering aims to improve robustness with respect to noise and ensure superpixel compactness \cite{wu_fuzzy_2021, ng_variational_2022}. Instead of growing regions another type of clustering is to apply a coarse-to-fine approach \cite{bergh_seeds_2013, yao_real-time_2015}. With this approach the method starts with a coarse region segmentation and swaps pixels based on an energy function. A major drawback of these methods is that the superpixel boundaries are set solely on local cues and there is no element that ensures that a dominant global image boundary is favoured to a local edge.

\begin{figure}
  \centering
  \hfill
    \begin{subfigure}{0.49\linewidth}
    \includegraphics[width=1.0\linewidth]{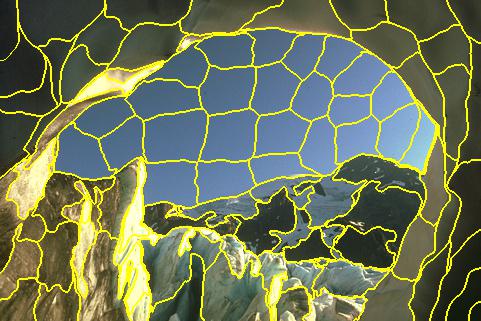}
    \caption{100 Superpixels}
    \label{fig:qual-intro-a}
  \end{subfigure}
  \hfill
  \begin{subfigure}{0.49\linewidth}
    \includegraphics[width=1.0\linewidth]{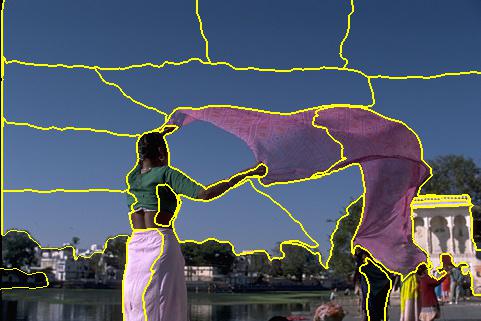}
    \caption{20 Superpixels}
    \label{fig:qual-intro-b}
  \end{subfigure}
\caption{Illustration of superpixel results generated by our proposed approach. Images are from BSDS500.}
\label{fig:intro}
\end{figure}

The recent success of deep learning based superpixel generators is due to the fact that the set of features extracted by neural networks can be much richer than the $labxy$ feature space that SLIC operates on. Some supervised methods rely on differentiable superpixel generation models \cite{ferrari_superpixel_2018, awaisu_fast_2019} and use backpropagation to optimize the deep feature generation with respect to a reconstruction and compactness loss. SEAL \cite{tu_learning_2018} is a loss that enables deep feature training with a non-differentiable graph-based superpixel generator. Alternatively, unsupervised models started to appear to avoid the need for costly annotations and extensive training. Zhu \etal \cite{zhu_learning_2021} formulated the superpixel generation as a lifelong learning task (LNS). Their method segments one image after another while at the same time optimizing the parameters of their method with respect to an unsupervised reconstruction and clustering loss. However, the authors warn about the lack of superpixel shape regularity in noisy backgrounds and trivial superpixels caused by the method's non-convergence. 

Recently, the implicit bias induced by the architectural design of neural networks has been used as prior for restoration tasks, such as denoising and inpainting \cite{ulyanov_deep_2018}, and is referred to as deep image prior (DIP). The success of DIP lies in the fact that when reconstructing an image from noise, the neural network first reproduces the low-frequency signal and only with a lag the high frequency part of the image. Several works have further investigated this behaviour and explored architectures with varying degrees of bias towards smooth signals \cite{heckel_denoising_2019}. Typically neural networks would still fit the noisy high-frequency signals after a certain amount of training time, however Heckel and Hand \cite{heckel_deep_2019} introduced an under-parameterized network to avoid this pitfall and is referred to as Deep Decoder.
Suzuki \etal \cite{suzuki_superpixel_2020} adapts the deep image prior (DIP) procedure with the aim to exploit better global features for superpixel generation. They add a regularized information maximization and a smoothness loss to the reconstruction loss to let a CNN directly output superpixel confidence maps. A pixel is then assigned to the superpixel with maximal confidence. Further refinements of this method like adding an edge-aware loss (EA) \cite{yu_edge-aware_2021} and a soft reconstruction loss \cite{eliasof_rethinking_2022} have been proposed. Even though these works showed impressive results compared to classical approaches, the optimization of the neural networks in these models require separate losses for enforcing smoothness and edge adherence. 



The work we present here differs from prior work in important ways. While optimizing convolutional kernels has been the default approach of most deep learning based superpixel generators, our algorithm is based on a non-convolutional decoder model that is mainly based on iterative channel combination and thresholding. We argue that this approach has two key advantages over existing methods. The first advantage is that by enforcing smoothness at the input of such a network, we can induce piece-wise smooth channels across all layers of our decoder including the output layer. This is because thresholding a smooth input, \ie, with a ReLU non-linearity, can only yield smooth contours and in limited amount. 
In combination with a reconstruction loss we find that our decoder sparsely reconstructs edges of the target image with a tendency to focus on smooth and globally significant contrasts. This is particularly useful, because unlike previous methods, we do not need to formulate nor balance smoothness and edge adherence losses. Instead, these are automatically enforced with the reconstruction loss through the smoothness of the input and the inductive bias of the non-convolutional deep decoder architecture. Besides the reconstruction loss we only rely on an additional positonal awareness loss to enforce superpixel compactness. The second advantage of non-convolutional decoder is that the activation maps from the last (hidden) layer, can be used as high quality pixel embeddings. 
As the network output is the sigmoid of a linear channel combination of these maps with dropout we can conclude that every edge in the reconstruction can be backtraced to at least one higher intensity contrast in them. This leads to strong edge adherence when clustering activation maps at the last layer for superpixels. Furthermore, we can smoothly blend spatial information into the pixel embeddings by extending the decoder's target image with positional encodings. Clustering these pixel embeddings achieves state-of-the-art superpixel generation results (see \Cref{fig:intro} for results of two challenging cases) on two natural image datasets as well as outperforming other methods in the downstream task of segmentation on microscopy images. To sum up, our contributions are three-fold:

\begin{itemize}[leftmargin=*]
	\setlength{\itemsep}{0pt}
	\setlength{\parsep}{-2pt}
	\setlength{\parskip}{-0pt}
	\setlength{\leftmargin}{-8pt}
	\vspace{-4pt}
\item We propose to incorporate smoothness and edge-sparsity in the design of the neural network set-up and thus omitting the need for explicit loss functions.
\item We exploit the piece-wise smooth activation maps enriched with spatial information as features for superpixel clustering.
\item Our method shows state-of-the art performance on two natural image datasets and on a downstream task of segmentation for microscopy images.
\end{itemize}

\section{Background}
\label{sec:back}

\subsection{Deep Decoder}

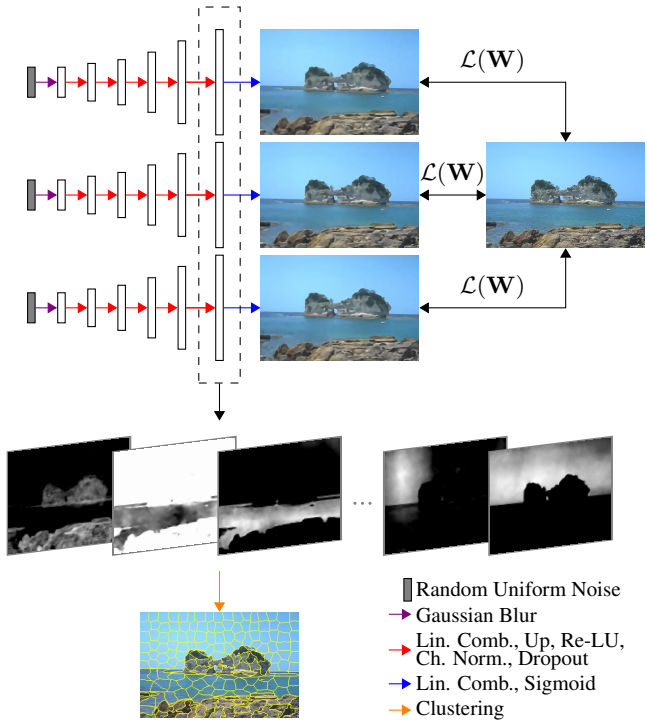
\begin{figure}[ht!]
    \centering
    \begin{tikzpicture}
        \node[inner sep=0pt] (orig) at (7.6,1.5){\includegraphics[height=1.4cm]{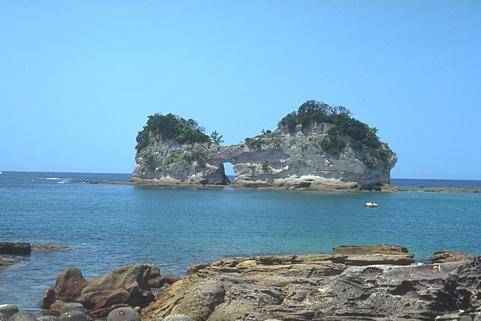}};
        
        \node[draw, minimum width=0.1cm, minimum height=0.4cm, inner sep=1pt, fill=gray] (noise) at (0.5,0){};
        \node[draw, minimum width=0.1cm, minimum height=0.4cm, inner sep=1pt] (Act0) at (0.9,0){};
        \node[draw, minimum width=0.1cm, minimum height=0.5cm, inner sep=1pt] (Act1) at (1.3,0){};
        \node[draw, minimum width=0.1cm, minimum height=0.6cm, inner sep=1pt] (Act2) at (1.7,0){};
        \node[draw, minimum width=0.1cm, minimum height=0.8cm, inner sep=1pt] (Act3) at (2.1,0){};
        \node[draw, minimum width=0.1cm, minimum height=1.1cm, inner sep=1pt] (Act4) at (2.5,0){};
        \node[draw, minimum width=0.1cm, minimum height=1.4cm, inner sep=1pt] (Act5) at (3.0,0){};
        \node[inner sep=0pt] (recon) at (4.6,0){\includegraphics[height=1.4cm]{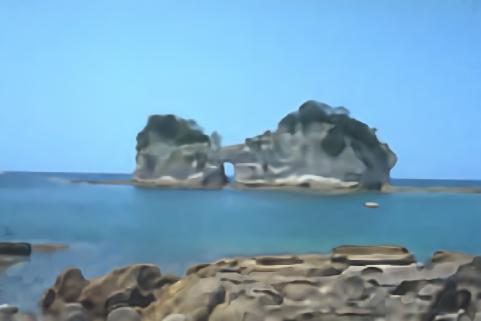}};

        \draw[-{Triangle}, violet] (noise) to (Act0);
        \draw[-{Triangle}, red] (Act0) to (Act1);
        \draw[-{Triangle}, red] (Act1) to (Act2);
        \draw[-{Triangle}, red] (Act2) to (Act3);
        \draw[-{Triangle}, red] (Act3) to (Act4);
        \draw[-{Triangle}, red] (Act4) to (Act5);
        \draw[-{Triangle}, red] (Act4) to (Act5);
        \draw[-{Triangle}, blue] (Act5) to (recon);
        \draw[{Triangle}-{Triangle}] (recon) to node[pos=0.5, above] {\small$\loss(\weights)$} (7.6,0) to (orig);

        \node[draw, minimum width=0.1cm, minimum height=0.4cm, inner sep=1pt, fill=gray] (noise) at (0.5,1.5){};
        \node[draw, minimum width=0.1cm, minimum height=0.4cm, inner sep=1pt] (Act0) at (0.9,1.5){};
        \node[draw, minimum width=0.1cm, minimum height=0.5cm, inner sep=1pt] (Act1) at (1.3,1.5){};
        \node[draw, minimum width=0.1cm, minimum height=0.6cm, inner sep=1pt] (Act2) at (1.7,1.5){};
        \node[draw, minimum width=0.1cm, minimum height=0.8cm, inner sep=1pt] (Act3) at (2.1,1.5){};
        \node[draw, minimum width=0.1cm, minimum height=1.1cm, inner sep=1pt] (Act4) at (2.5,1.5){};
        \node[draw, minimum width=0.1cm, minimum height=1.4cm, inner sep=1pt] (Act5) at (3.0,1.5){};
        \node[inner sep=0pt] (recon) at (4.6,1.5){\includegraphics[height=1.4cm]{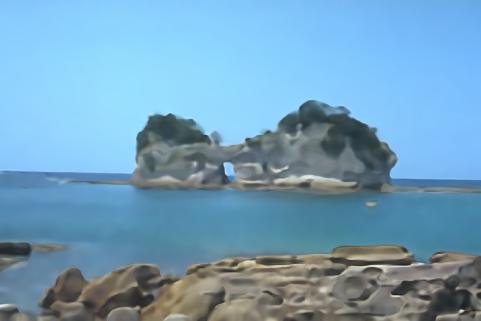}};

        \draw[-{Triangle}, violet] (noise) to (Act0);
        \draw[-{Triangle}, red] (Act0) to (Act1);
        \draw[-{Triangle}, red] (Act1) to (Act2);
        \draw[-{Triangle}, red] (Act2) to (Act3);
        \draw[-{Triangle}, red] (Act3) to (Act4);
        \draw[-{Triangle}, red] (Act4) to (Act5);
        \draw[-{Triangle}, red] (Act4) to (Act5);
        \draw[-{Triangle}, blue] (Act5) to (recon);
        \draw[{Triangle}-{Triangle}] (recon) to node[pos=0.5, above] {\small$\loss(\weights)$} (orig);

        \node[draw, minimum width=0.1cm, minimum height=0.4cm, inner sep=1pt, fill=gray] (noise) at (0.5,3){};
        \node[draw, minimum width=0.1cm, minimum height=0.4cm, inner sep=1pt] (Act0) at (0.9,3){};
        \node[draw, minimum width=0.1cm, minimum height=0.5cm, inner sep=1pt] (Act1) at (1.3,3){};
        \node[draw, minimum width=0.1cm, minimum height=0.6cm, inner sep=1pt] (Act2) at (1.7,3){};
        \node[draw, minimum width=0.1cm, minimum height=0.8cm, inner sep=1pt] (Act3) at (2.1,3){};
        \node[draw, minimum width=0.1cm, minimum height=1.1cm, inner sep=1pt] (Act4) at (2.5,3){};
        \node[draw, minimum width=0.1cm, minimum height=1.4cm, inner sep=1pt] (Act5) at (3.0,3){};
        \node[inner sep=0pt] (recon) at (4.6,3){\includegraphics[height=1.4cm]{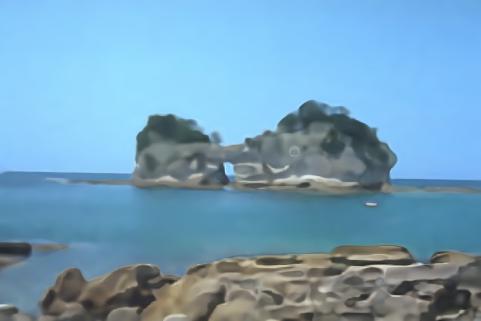}};

        \draw[-{Triangle}, violet] (noise) to (Act0);
        \draw[-{Triangle}, red] (Act0) to (Act1);
        \draw[-{Triangle}, red] (Act1) to (Act2);
        \draw[-{Triangle}, red] (Act2) to (Act3);
        \draw[-{Triangle}, red] (Act3) to (Act4);
        \draw[-{Triangle}, red] (Act4) to (Act5);
        \draw[-{Triangle}, red] (Act4) to (Act5);
        \draw[-{Triangle}, blue] (Act5) to (recon);
        \draw[{Triangle}-{Triangle}] (recon) to node[pos=0.5, above] {\small$\loss(\weights)$} (7.6,3) to (orig);

        \node[draw,dashed, minimum height=5cm, minimum width= 0.55cm] (actmaps) at (3,1.5) {};
        
        \draw[-{Triangle}] (actmaps) to (3,-1.5);

        \begin{scope}[cm={0.8,0.1,0,1,(1,-2.5)}]
            \node[transform shape, draw, color=gray, ultra thick, inner sep=0mm] {\includegraphics[width=2cm]{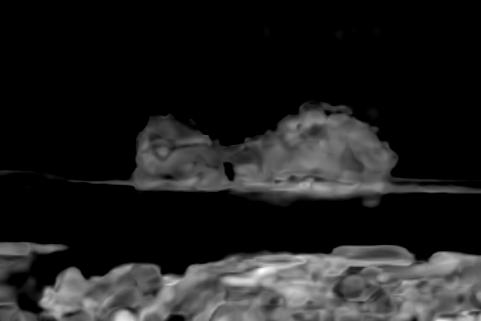}};
        \end{scope}
        \begin{scope}[cm={0.8,0.1,0,1,(2.4,-2.5)}]
            \node[transform shape, draw, color=gray, ultra thick,inner sep=0mm] {\includegraphics[width=2cm]{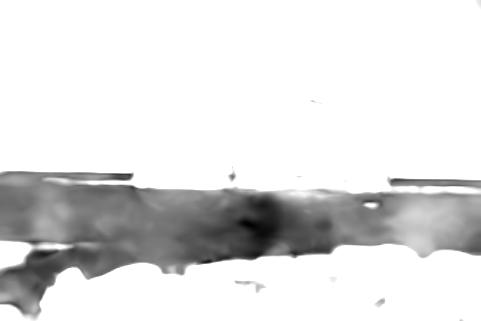}};
        \end{scope}
        \begin{scope}[cm={0.8,0.1,0,1,(3.8,-2.5)}]
            \node[transform shape, draw, color=gray, ultra thick,inner sep=0mm] {\includegraphics[width=2cm]{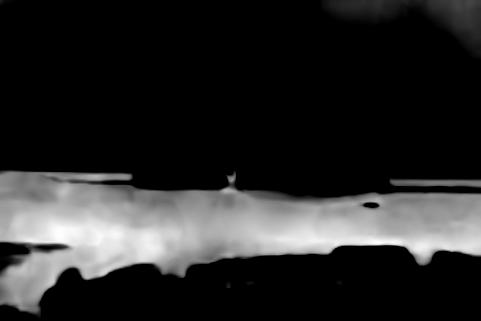}};
        \end{scope}
        \fill[fill=gray](4.8,-2.6) circle (0.02);
        \fill[fill=gray](4.9,-2.6) circle (0.02);
        \fill[fill=gray](5.0,-2.6) circle (0.02);
        \begin{scope}[cm={0.8,0.1,0,1,(6,-2.5)}]
            \node[transform shape, draw, color=gray, ultra thick,inner sep=0mm] {\includegraphics[width=2cm]{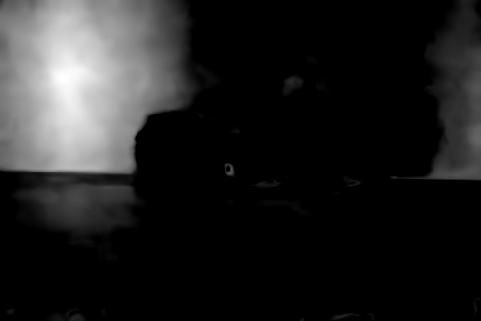}};
        \end{scope}
        \begin{scope}[cm={0.8,0.1,0,1,(7.4,-2.5)}]
            \node[transform shape, draw, color=gray, ultra thick,inner sep=0mm] {\includegraphics[width=2cm]{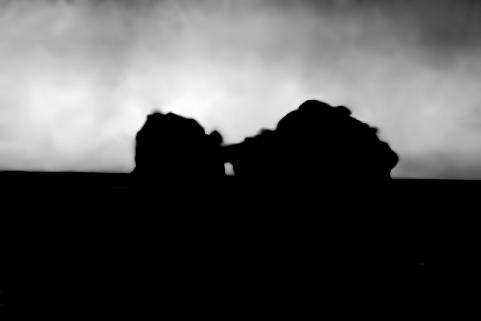}};
        \end{scope}

        \node[inner sep=0pt] (sp) at (3,-4.75){\includegraphics[height=1.4cm]{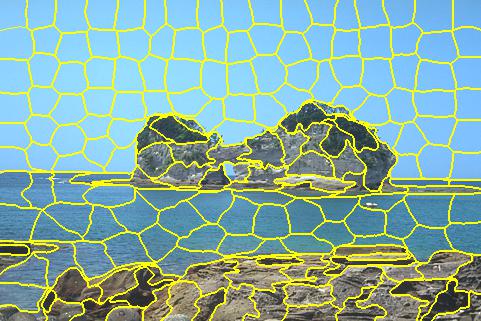}};
        \draw[-{Triangle}, color=orange] (3,-3.5) to (sp);

        %

        \node[draw, minimum width=0.1cm, minimum height=0.3cm, inner sep=1pt, fill=gray] (noise) at (5.5,-3.75){};
        \draw[-{Triangle}, violet] (5.25,-4.075) to (5.55, -4.075);
        \draw[-{Triangle}, red] (5.25,-4.55) to (5.55, -4.55);
        \draw[-{Triangle}, blue] (5.25,-5) to (5.55, -5);
        \draw[-{Triangle}, orange] (5.25,-5.35) to (5.55, -5.35);
        \node[align=left, inner sep=0pt, font=\footnotesize] (dm_leg) at (7,-4.7) {Random Uniform Noise\\
                                                                Gaussian Blur\\
                                                                Lin. Comb., Up, Re-LU,\vspace{-0.15cm}\\
                                                                Ch. Norm., Dropout\\
                                                                Lin. Comb., Sigmoid\\
                                                                Clustering\\};

    \end{tikzpicture}
    \caption{\textbf{Method Overview.} Blurred random fields serve as input for 3 Deep Decoders consisting of 5 consecutive blocks of linear channel combinations (Lin-Comb.), bi-linear up-sampling (Up), rectified linear units (Re-LU), channel-wise normalisation and dropout. This is followed by by a another linear channel combination and a sigmoid operation. The weights of the 3 deep decoders are optimized with respect to loss $\loss$. After that pixel embeddings are generated from the final layers' activation maps and clustered for a superpixel partition.}
    \label{fig:overview}
\end{figure}

Our network is based on the deep decoder structure introduced by Heckel and Hand \cite{heckel_deep_2019}. A deep decoder $\decoder$ is comprised of blocks consisting of linear channel combinations (also referred to as 1$\times$1 convolutions), bi-linear up-sampling, a rectified linear unit and a batch normalisation operation. These blocks are followed by another linear channel combination and a sigmoid activation, as it is visualised in \Cref{fig:overview}. The number of channels $\numchannels$ is constant for all layers except for the last one where it is reduced to the target's channel size, \eg, $\targetchan=3$ for RGB images. The concatenation of all weights from all layers $\weights$ spans a class of functions that map a low dimensional input tensor $\inimg$ to an $\targetchan$-channel  image $\decoder(\weights; \inimg) \in [0,1]^{\targetchan \times \width \times \height}$ of width $\width$ and height $\height$ . Just like the DIP \cite{ulyanov_deep_2018} the deep decoder is fitted onto a target image $\targetimage$ by optimizing its weights with respect to a mean squared error reconstruction loss $\loss(\weights)$:
\begin{align}\small
  \hat{\weights} &= \argmin_\weights \loss(\weights, \targetimage; \inimg)\label{eq:minigeneral}\\
  &= \argmin_\weights \norm{\decoder(\weights; \inimg) - \targetimage}^2.
  \label{eq:minidd}
\end{align}
Because the Deep Decoder is drastically under-parameterized, it can only capture a fraction of the original image's complexity. It has been shown that this allows for stable convergence and state-of-the-art performance in image denoising, superresolution and impainting tasks \cite{heckel_deep_2019, heckel_denoising_2019}. The idea of using untrained neural networks for unsupervised segmentation has been introduced with the Double-DIP method \cite{gandelsman_double-dip_2019} where two competing DIPs are used to differentiate foreground from background based on the assumption that they follow different image statistics. Unlike this prior work, we do not let two separate priors disentangle mixed image statistics but use the fact that each channel of one decoder tends to focus on the reconstruction of spatially connected pixels following a low variance distribution.
 
\subsection{Positional Encoding}
\label{sec:posenc}
It was found that neural networks can have problems reconstructing high frequency components when directly operating on the $xy$ space. In the context of neural radiance fields for view synthesis \cite{mildenhall_nerf_2021} Fourier mappings have been introduced as positional encoding. This was further investigated by Tancik \etal \cite{tancik_fourier_2020}. The original encoding $\ogposenc^{\posfreq}_{\hat{z}}$ is defined by a normalized coordinate $\hat{z} \in \{\hat{x}, \hat{y}\} \subset [0,1]^{\width \times \height}$ and a parameter $\posfreq \in \integers^{+}$ that controls the frequency range of the encoding:
\begin{align}\small
  \ogposenc^{\posfreq}_{\hat{z}} = (\sin(2^0\pi\hat{z}), \cos(2^0\pi\hat{z}), \dots, \sin(2^l\pi\hat{z}), \cos(2^l\pi\hat{z})).
\end{align}

\subsection{SLIC}
SLIC \cite{achanta_slic_2012} clusters pixels in the $labxy$ space with respect to a weighted distance measure. It starts by initializing clusters with pixels from an evenly spaced grid over the image. Each of these initial pixels then gets replaced with the minimum gradient pixel from its immediate neighbourhood. The cluster initialization is followed by repeated assignment and update steps. In the assignment SLIC assigns pixels to the best-matching cluster in their neighbourhood according to the weighted distance measure to the cluster's center. In the update step the clusters' centers are updated with their average pixel feature. After the convergence criterion is met, spatial connectivity is enforced by assigning the smallest surplus connected components to their largest neighbouring connected component.

\section{Our Approach}
\label{sec:method}

Our strategy is to exploit the inductive bias of an under-parameterized and non-convolutional neural network so that when reconstructing an image from a smooth input, the network is forced to only capture prominent high-frequency details in the image to optimally reconstruct it, which in turn yields pixel embeddings that can be enriched with spatial information and that we can utilize for superpixel generation. In this section, we explain how our algorithm concentrates edge (\cref{sec:edge_sparse}) and spatial information (\cref{sec:pos-aware}) in a decoder's last layer. Then, we use it to extract informative pixel features (\cref{sec:featureextract}) and eventually cluster them into high quality superpixels (\cref{sec:superpixelgen}).
An overview of our approach can be seen in \Cref{fig:overview}.



\subsection{Edge Sparsity in a Deep Decoder}
\label{sec:edge_sparse}

We base our model architecture on the Deep Decoder, which besides the Re-LU operation is only composed of smoothing (bi-linear up-sampling) and smoothness preserving operations (linear channel combination, batch normalization). By providing a smooth input $\inimg*\gaussiankernel$ to a Deep Decoder, obtained by blurring random uniform noise $\inimg$ sampled from the interval $[-1,1]$ with a Gaussian kernel $K$ of standard deviation $\blurstd$, we introduce a strong neighbouring pixel coupling. So the Re-LU is the only element that can introduce additional non-linearities and edges going from one layer to the next. The Re-LU splits each channel into activated and non-activated regions by thresholding and on each region's circumference an edge is reproduced. This on the one hand enforces smooth and, especially, closed connected edges that are particularly valuable for the subsequent superpixel segmentation. On the other hand this points to a strong link between the number of edges introduced to the number of activated regions in each channel, which allows us to do further investigation into the edge sparsity dynamic of decoding blurry input.

The expected number $\mathbb{E}[{\it EC}]$ of activated regions of a $Z$-thresholded random uniform image blurred with a kernel was first formulated by Worsley \etal \cite{worsley_three-dimensional_1992}. With a Gaussian kernel of standard deviation $\blurstd$ on an $N$-pixel image it evaluates to
\begin{align}\small
    \mathbb{E}[{\it EC}] = \frac{N^2}{2\blurstd^2}(2\pi)^{-\frac{3}{2}}Ze^{-\frac{1}{2}Z^2}.
\label{eq:eulercharacteristic}
\end{align}

\begin{figure}[t]
    \centering
    \begin{subfigure}{0.48\linewidth}
        \begin{tikzpicture}
            \begin{axis}[
            axis lines = left,
            xlabel = {$\blurstd$},
            ylabel = {\#$P$},
            xmin = 1,
            xmax = 59,
            ymin = 0,
            ymax = 40,
            width=\linewidth,
            height=0.9\linewidth,
            xlabel style={xshift=1.2cm, yshift=0.55cm},
            ytick = {0,10,20,30},
            ylabel style={rotate=-90, xshift=0.8cm, yshift=0.9cm}
            ]
            \addplot table [x=blur, y=avg_partitions, col sep=comma]{figures/partition_blur/blur_partition.csv};
            \end{axis}
        \end{tikzpicture}
        \caption{Average number of partitions}
        \label{fig:blur_edges-a}
    \end{subfigure}
    \hfill
    \begin{subfigure}{0.48\linewidth}
        \includegraphics[width=1.0\linewidth]{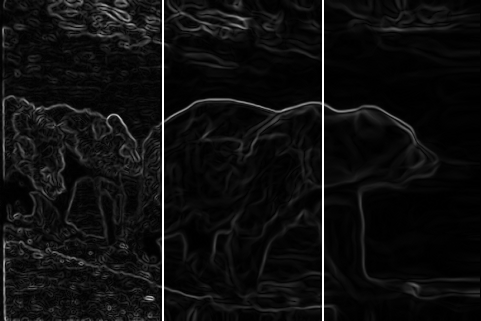}
        \caption{Sparse Edge Gradients}
        \label{fig:blur_edges-b}
    \end{subfigure}
    \begin{subfigure}{0.48\linewidth}
        \includegraphics[width=1.0\linewidth]{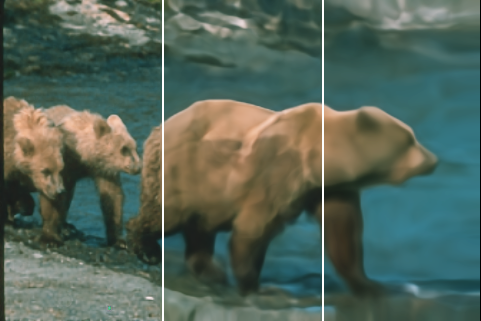}
        \caption{Sparse Edge Reconstructions}
        \label{fig:blur_edges-c}
    \end{subfigure}
    \hfill
    \begin{subfigure}{0.48\linewidth}
        \includegraphics[width=1.0\linewidth]{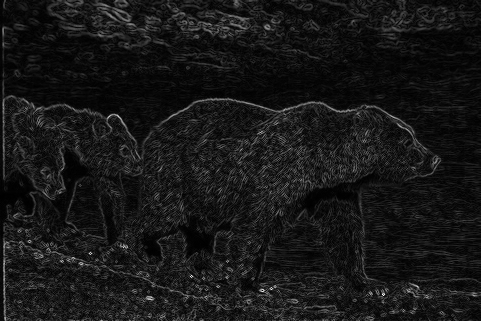}
        \caption{Original Gradients}
        \label{fig:blur_edges-d}
    \end{subfigure}
    \caption{\textbf{Illustration of edge sparsity introduced by deep decoder.} We decode the same noise blurred at different levels to an image of the BSDS500. In \Cref{fig:blur_edges-a} we plot the average number of Re-LU partitions \#$P$ in the last layer's channels of the fitted deep decoder against the standard deviation $\blurstd$ of the Gaussian kernel used to blur the input. In \Cref{fig:blur_edges-c} and \Cref{fig:blur_edges-b} we display the image reconstructions and corresponding gradients with $\blurstd=1$, $\blurstd=21$, $\blurstd=41$, respectively. For reference we provided the original image's gradient in \Cref{fig:blur_edges-d}.}
    \label{fig:blur_edges}
\end{figure}
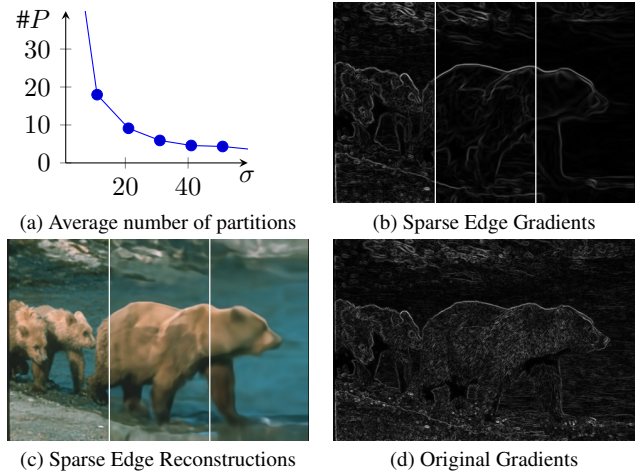

From  Eq.~\ref{eq:eulercharacteristic} we infer an inverse square relationship between the expected number of activated regions and the kernel's standard deviation $\blurstd$. While the assumption of Gaussian blurred uniform random input only holds for the first layer, we expect a similar relationship in the last layer because previous layers are likely to only add a limited amount of edges and also because the up-sampling operation in every layer further smooths each channel before thresholding. Experimental results displayed in \Cref{fig:blur_edges-a} support this claim. By our preceding argument this strongly indicates a similar correlation between the number of edges reproduced by the deep decoder and the Gaussian kernel's standard deviation $\blurstd$. We provide a qualitative example of how the number of reconstructed edges changes with a growing Gaussian kernel in \Cref{fig:blur_edges}. We can observe how the network with a high input blur focuses on edges corresponding to prominent boundaries between the bear and the water rather than arbitrary ones within the fur. Superpixel generation can greatly profit from this main edge prioritization.

\subsection{Positional Awareness}
\label{sec:pos-aware}

Positional awareness ensures that pixels which are close have a higher likelihood of being clustered together and those far away are assigned to different clusters. One trivial way of incorporating this would be to add the x-y coordinates as pixel-features and to cluster accordingly. We have found this to be a very strong constraint on clustering, which often ignored other pixel-features. Thus we propose to use positional encoding for the Deep Decoder by not only reconstructing the original image but also modulated versions of the original image. This way we can encode spatial information into the activation maps.
To generate these masked versions we use sinusoidal encodings $\posenc^{\posfreq}_{\hat{z}}$ similar to the encodings in \Cref{sec:posenc}. The random offset $\offset$ is an independent uniform sample from $[0,2\pi]$:
\begin{align}\small
  \posenc^{\posfreq}_{\hat{z}} = \frac{1}{2}  ( 
  &\sin(2^1 \pi \hat{z} + \offset_1), \quad -\sin(2^1 \pi \hat{z} + \offset_1), \nonumber\\ 
  &\cos(2^1 \pi \hat{z} + \offset_1), \quad -\cos(2^1 \pi \hat{z} + \offset_1), \nonumber\\ 
  &\dots \nonumber\\ 
  &\sin(2^l \pi \hat{z} + \offset_l), \quad -\sin(2^l \pi \hat{z} + \offset_l), \nonumber\\ 
  &\cos(2^l \pi \hat{z} + \offset_l), \quad -\cos(2^l \pi \hat{z} + \offset_l)) + \frac{1}{2}.
  \label{eq:posenc}
\end{align}

By concatenating the sinusoidal positional encodings in both directions we finally get $\posenc^{\posfreq} = (\posenc^{\posfreq}_{\hat{x}}, \posenc^{\posfreq}_{\hat{y}}) \in [0,1]^{8\posfreq \times \width \times \height}$. Instead of letting the Deep Decoder reconstruct these positional encodings directly we concatenate the element-wise products of the encodings and the lightness channel $\lightness$ of the CIELAB color space to the target image $\targetimage$ that the Deep Decoder is fitted on. This way the Deep Decoder's ability to mirror image contrasts with Re-LU cuts does not get distorted.
\begin{equation}\small
  \targetimage = \left( \img_\text{R}, \img_\text{G}, \img_\text{B}, \posenc^{\posfreq}_1 \odot\lightness, \dots, \posenc^{\posfreq}_{8\posfreq}\odot\lightness\right).
  \label{eq:targetimg}
\end{equation}

To control the compactness of the resulting superpixels we introduce a compactness parameter $\lossweight$ which determines the importance of spatial proximity for superpixel clustering and can be set according to the properties required by a downstream task.
\begin{equation}\small
  \loss = (1-\lossweight) \reconloss + \lossweight \spatialloss.
  \label{eq:loss}
\end{equation}
The first term is a classic reconstruction loss $\reconloss$ with respect to the original RGB channels:
\begin{equation}\small
  \reconloss(\weights, \targetimage) = \frac{1}{\width\height} \sum_{i=1}^3 \norm{\decoder(\weights; \inimg * \gaussiankernel)-\targetimage^i}^2.
  \label{eq:reconloss}
\end{equation}
The spatial awareness loss $\spatialloss$ comprises the second loss and is the mean squared error loss with respect to the positionally encoded lightness channels:
\begin{equation}\small
  \!\!\spatialloss(\weights, \targetimage) \!=\! \frac{1}{\width\height} \sum_{i=3}^{8\posfreq + 3}\ \norm{\decoder(\weights; \inimg * \gaussiankernel)-\targetimage^i}^2.\!\!
  \label{eq:spatialloss}
\end{equation}


\subsection{Enforcing Informative Feature Layers}\label{sec:featureextract}
To enable meaningful feature extraction from the activation maps in the last layer there should not be degenerate channels which do not contribute to the final reconstruction. This could happen by having an empty feature map or artifacts unrelated to the original image could occur in the features maps which cancel each other out in the last linear channel combination. To avoid these two scenarios we implement a dropout function after each layer of the deep decoder. The dropout's averaging behaviour first makes sure that every activation map contains meaningful information about the original image and also prevents the above mentioned artifacts. After an initial convergence we can disable the dropout without losing these two favourable effects. We do this with the intention of allowing the decoder to introduce more detailed edge and spatial information in the activation maps. As it can be seen in \Cref{fig:overview} the dropout will enforce important edge information with respect to the reconstruction error in multiple activation maps which will give more weight to globally significant edges in the clustering step.  


In addition, Heckel and Hand \cite{heckel_deep_2019} have already shown that the performance of Deep Decoders can be significantly improved by fitting multiple randomly initialized Deep Decoders in a row on the same image and then averaging the reconstructions. We follow this approach and fit $\ddnum$ Deep Decoders and concatenate the activation maps of the last layer into $\chmaps \in \real^{\ddnum\numchannels\times\width\times\height}$.

\subsection{Superpixel Generation}\label{sec:superpixelgen}

The pixel features (or embeddings) $\pixelfeat(x,y)$ that will be used for superpixel generation are extracted from the activation maps $\chmaps$ of the last layer by concatenating the intensity values at each pixel location:
\begin{equation}\small
  \pixelfeat(x,y) = \begin{pmatrix} (\chmaps)_{1,x,y} \\
                    \vdots \\
                    (\chmaps)_{\ddnum\numchannels,x,y}
                \end{pmatrix}
                \in \real^{\ddnum\numchannels}.
  \label{eq:pixelfeat}
\end{equation}

For the actual segmentation we use SLIC-like clustering approach. We initialize the cluster centers in a uniform grid over the image and select the pixels with the lowest average channel gradient in a $5\times5$ neighbourhood. Then we iteratively apply an assignment, an update step with an unweighted Euclidean distance measure. Connectivity is enforced after every update step by performing a connected component analysis for each cluster and keeping the biggest shape while the other connected components (if there are any) get randomly assigned to one of their neighbours. We repeat this 100 times or until the number of changed pixel labels from one to the next step is below a threshold.


\section{Experiments}
We conduct experiments on two natural image datasets and compare the superpixel quality quantitatively and qualitatively with current state-of-the-art methods. The first dataset we used is the BSDS500 dataset \cite{arbelaez_contour_2011} providing 500 natural images with multiple human segmentations. Because this dataset was designed to test edge detectors and superpixel algorithms, we include results on a second more downstream oriented dataset. The PASCAL-Context dataset \cite{mottaghi_role_2014} contains natural images with more than 400 contextual labels. Out of computational considerations we randomly sampled 200 images from the PASCAL-Context dataset. We compared our approach with earlier mentioned methods SLIC, SNIC, edge-aware untrained neural networks (EA) and the lifelong-learning non-iterative network (LNS).

Finally, we show our method's value by using it in an end-to-end foreground segmentation pipeline on the DeepVess dataset by Haft-Javaherian \etal \cite{haft-javaherian_deep_2019}. The dataset consists of a $256 \times 256 \times 200$ multi-photon image volume of in-vivo brain vasculature and was published to compare vessel segmentation methods. We start by generating 600 superpixels on each $256 \times 256$ slice with EA, SNIC and our method. To account for different base intensities and multiplicative noise across the slice we assign each superpixel $i$ its weber coefficient $w_i$,
\begin{equation}\small
  w_i = \frac{\text{Superpixel Intensity}-\text{Neighbourhood Intensity}}{\text{Neighbourhood Intensity}}.
  \label{eq:weber}
\end{equation}
On the resulting weber maps we apply classic Li-Thresholding \cite{li_minimum_1993} to generate a foreground estimate. We compare the segmentation results with Li-Thresholding on the original image and on the BM3D-denoised version \cite{dabov2007image}, as well as a current state-of-the-art supervised CNN-Network called DeepVess \cite{haft-javaherian_deep_2019} and a second human annotation. Furthermore, we include Double-DIP \cite{gandelsman_double-dip_2019} results that were intialized with Li-Thresholded original images. Experiments, presented in the supplementary material, have shown that much smoother foreground masks can be extracted by decreasing the deep decoder's number of layers and channels to fully exploit the its denoising ability. However, to ensure comparability with other superpixel algorithms we used the same hyperparameters for all tasks in the result section.

\subsection{Implementation Details}
During the deep decoder fitting we disable the dropout and reduce the learning by a factor of 0.8 after 1000 optimization steps. We chose the compactness parameter to be $\lossweight=0.1$. Each of the 3 deep decoders was fitted with an ADAM optimizer for 1500 steps. In order to set the input blur level independent of the target image size we introduce a blur factor and set it to $\blurfactor = 0.0001$ with which we can calculate the Gaussian kernel's standard deviation $\blurstd = 2\lfloor\frac{b\width\height}{2} \rfloor+1$. As discussed in \Cref{sec:ablation} these values were chosen based on experiments on a subset of the BSDS500's train images. We use the same hyperparamters for our experiments on the BSDS500, Pascal-Context and the DeepVess dataset. With these parameters it takes on average 39.2 seconds to fit a Deep Decoder to an image of the BSDS500 test set on a NVIDIA TITAN Xp.

\subsection{Evaluation Metrics}
We base our quantitative comparison on four metrics. The first three metric definitions were defined by Liu \etal \cite{liu_entropy_2011}. The undersegmentation error (USE) measures the amount of leakage around ground truth boundaries. The boundary recall (BR) is the percentage of ground truth boundary pixels with a superpixel boundary in a $2$-pixel neighbourhood. The achievable segementation accuracy (ASA) is highly correlated with the USE but included for comparability. It penalizes superpixels that cover more than one ground truth segment. Furthermore, we want to quantitatively assess the compactness of generated superpixels. We used the compactness metric that was introduced by Schick \etal \cite{schick_measuring_2012} as regular shaped superpixels are not only visually more appealing but indicate spatial coherence and facilitate further processing.
As opposed to our method, most methods do not permit strict control over the number of superpixels, as they do not enforce superpixel connectivity. To allow for a fair comparison we first sample superpixels at different cluster parameters, \ie, the number of desired superpixels a superpixel algorithm gets as an input. We then determine the actual number of superpixels for each segmentation with a connected component analysis and finally log the evaluation metrics accordingly. For the BSDS500 the metrics are averaged over each human annotation, as we want to measure how well the same set of superpixels fits into different ground truth segmentations.

For evaluation of our predicted masks in the vessel segmentation task we calculated the sensitivity $S_e$, specificity $S_p$, the Jaccard-Index $JI$ and the Dice Coefficient $DC$ with respect to the dataset's gold standard. They are all based on the number of true positive $T_P$, true negative $T_N$, false positive $F_P$ and false negative $F_N$ pixels.
\begin{align*}\small
    S_e &=  \frac{T_P}{T_P+F_N}, \quad &S_p &= \frac{T_N}{T_N+F_P}\\
    DC &= \frac{2T_P}{2T_P+F_P+F_N}, \quad &JI &= \frac{T_P}{T_P+F_P+F_N}
\end{align*}
Haft-Javaherian \etal \cite{haft-javaherian_deep_2019} stresses the significance of $DC$ and $JI$ as performance indicators for vessel segmentation as they focus more on the detection accuracy of the foreground than on the background. Moreover, they are better suited for segmentation tasks where the foreground covers much less area than the background, as it is the case with most microscopic images of cells and vessels.

\subsection{Results}

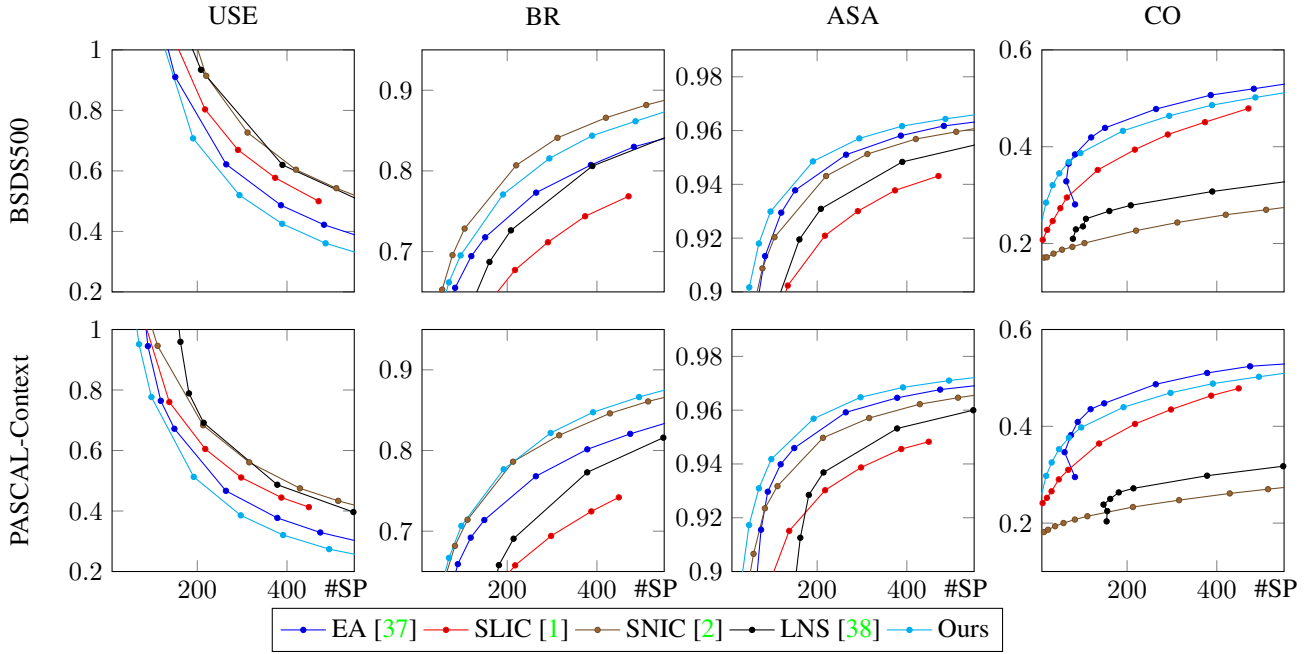
\begin{figure*}
    \centering
  \begin{tikzpicture}
    \begin{groupplot}[
      group style={group size=4 by 2, x descriptions at=edge bottom, vertical sep=0.5cm, horizontal sep = 0.9cm},
      width=4.8cm, height=4.8cm
    ]
    \nextgroupplot[title=USE, ymin = 0.2, ymax= 1.0, xmin=10, xmax=550, ylabel=BSDS500]
      \addplot+[mark = *, mark size=1pt] table[x=num_regions, y=USE, col sep=comma] {figures/sp_metrics/BSDS_ea.csv};
      \addplot+[mark = *, mark size=1pt] table[x=num_regions, y=USE, col sep=comma] {figures/sp_metrics/BSDS_slic.csv};
      \addplot+[mark = *, mark size=1pt] table[x=num_regions, y=USE, col sep=comma] {figures/sp_metrics/BSDS_snic.csv};
      \addplot+[mark = *, mark size=1pt] table[x=num_regions, y=USE, col sep=comma] {figures/sp_metrics/BSDS_lns.csv};
      \addplot+[mark = *, mark size=1pt, color=cyan, mark options={fill=cyan}] table[x=num_regions, y=USE, col sep=comma] {figures/sp_metrics/BSDS_ddseg.csv};
    \hfill
    \nextgroupplot[title=BR, ymin = 0.65, ymax= 0.95, xmin=10, xmax=550]
      \addplot+[mark = *, mark size=1pt] table[x=num_regions, y=BR, col sep=comma] {figures/sp_metrics/BSDS_ea.csv};
      \addplot+[mark = *, mark size=1pt] table[x=num_regions, y=BR, col sep=comma] {figures/sp_metrics/BSDS_slic.csv};
      \addplot+[mark = *, mark size=1pt] table[x=num_regions, y=BR, col sep=comma] {figures/sp_metrics/BSDS_snic.csv};
      \addplot+[mark = *, mark size=1pt] table[x=num_regions, y=BR, col sep=comma] {figures/sp_metrics/BSDS_lns.csv};
      \addplot+[mark = *, mark size=1pt, color=cyan, mark options={fill=cyan}] table[x=num_regions, y=BR, col sep=comma] {figures/sp_metrics/BSDS_ddseg.csv};
    \hfill
    \nextgroupplot[title=ASA, ymin = 0.9, ymax= 0.99, xmin=10, xmax=550]
      \addplot+[mark = *, mark size=1pt] table[x=num_regions, y=ASA, col sep=comma] {figures/sp_metrics/BSDS_ea.csv};
      \addplot+[mark = *, mark size=1pt] table[x=num_regions, y=ASA, col sep=comma] {figures/sp_metrics/BSDS_slic.csv};
      \addplot+[mark = *, mark size=1pt] table[x=num_regions, y=ASA, col sep=comma] {figures/sp_metrics/BSDS_snic.csv};
      \addplot+[mark = *, mark size=1pt] table[x=num_regions, y=ASA, col sep=comma] {figures/sp_metrics/BSDS_lns.csv};
      \addplot+[mark = *, mark size=1pt, color=cyan, mark options={fill=cyan}] table[x=num_regions, y=ASA, col sep=comma] {figures/sp_metrics/BSDS_ddseg.csv};
    \hfill
    \nextgroupplot[title=CO, ymin = 0.1, ymax= 0.6, xmin=10, xmax=550]
      \addplot+[mark = *, mark size=1pt] table[x=num_regions, y=CO, col sep=comma] {figures/sp_metrics/BSDS_ea.csv};
      \addplot+[mark = *, mark size=1pt] table[x=num_regions, y=CO, col sep=comma] {figures/sp_metrics/BSDS_slic.csv};
      \addplot+[mark = *, mark size=1pt] table[x=num_regions, y=CO, col sep=comma] {figures/sp_metrics/BSDS_snic.csv};
      \addplot+[mark = *, mark size=1pt] table[x=num_regions, y=CO, col sep=comma] {figures/sp_metrics/BSDS_lns.csv};
      \addplot+[mark = *, mark size=1pt, color=cyan, mark options={fill=cyan}] table[x=num_regions, y=CO, col sep=comma] {figures/sp_metrics/BSDS_ddseg.csv};
    \hfill
    \nextgroupplot[ymin = 0.2, ymax= 1.0, xmin=10, xmax=550, ylabel={PASCAL-Context}, xlabel={\#SP},
    xlabel style={xshift=1.5cm, yshift=0.55cm}]
      \addplot+[mark = *, mark size=1pt] table[x=num_regions, y=USE, col sep=comma] {figures/sp_metrics/VOC_Context_ea.csv};
      \addplot+[mark = *, mark size=1pt] table[x=num_regions, y=USE, col sep=comma] {figures/sp_metrics/VOC_Context_slic.csv};
      \addplot+[mark = *, mark size=1pt] table[x=num_regions, y=USE, col sep=comma] {figures/sp_metrics/VOC_Context_snic.csv};
      \addplot+[mark = *, mark size=1pt] table[x=num_regions, y=USE, col sep=comma] {figures/sp_metrics/VOC_Context_lns.csv};
      \addplot+[mark = *, mark size=1pt, color=cyan, mark options={fill=cyan}] table[x=num_regions, y=USE, col sep=comma] {figures/sp_metrics/VOC_Context_ddseg.csv};
    \hfill
    \nextgroupplot[ymin = 0.65, ymax= 0.95, xmin=10, xmax=550, xlabel={\#SP},
    xlabel style={xshift=1.5cm, yshift=0.55cm}]
      \addplot+[mark = *, mark size=1pt] table[x=num_regions, y=BR, col sep=comma] {figures/sp_metrics/VOC_Context_ea.csv};
      \addplot+[mark = *, mark size=1pt] table[x=num_regions, y=BR, col sep=comma] {figures/sp_metrics/VOC_Context_slic.csv};
      \addplot+[mark = *, mark size=1pt] table[x=num_regions, y=BR, col sep=comma] {figures/sp_metrics/VOC_Context_snic.csv};
      \addplot+[mark = *, mark size=1pt] table[x=num_regions, y=BR, col sep=comma] {figures/sp_metrics/VOC_Context_lns.csv};
      \addplot+[mark = *, mark size=1pt, color=cyan, mark options={fill=cyan}] table[x=num_regions, y=BR, col sep=comma] {figures/sp_metrics/VOC_Context_ddseg.csv};
    \hfill
    \nextgroupplot[ymin = 0.9, ymax= 0.99, xmin=10, xmax=550, legend style={legend columns=6,at={(-1.9,-0.15)},anchor=north west}, xlabel={\#SP},
    xlabel style={xshift=1.5cm, yshift=0.55cm}]
      \addplot+[mark = *, mark size=1pt] table[x=num_regions, y=ASA, col sep=comma] {figures/sp_metrics/VOC_Context_ea.csv};
      \addplot+[mark = *, mark size=1pt] table[x=num_regions, y=ASA, col sep=comma] {figures/sp_metrics/VOC_Context_slic.csv};
      \addplot+[mark = *, mark size=1pt] table[x=num_regions, y=ASA, col sep=comma] {figures/sp_metrics/VOC_Context_snic.csv};
      \addplot+[mark = *, mark size=1pt] table[x=num_regions, y=ASA, col sep=comma] {figures/sp_metrics/VOC_Context_lns.csv};
      \addplot+[mark = *, mark size=1pt, color=cyan, mark options={fill=cyan}] table[x=num_regions, y=ASA, col sep=comma] {figures/sp_metrics/VOC_Context_ddseg.csv};
      \legend{EA \cite{yu_edge-aware_2021}, SLIC \cite{achanta_slic_2012}, SNIC \cite{achanta_superpixels_2017},LNS \cite{zhu_learning_2021}, Ours}
    \hfill
    \nextgroupplot[ymin = 0.1, ymax= 0.6, xmin=10, xmax=550, xlabel={\#SP},
    xlabel style={xshift=1.5cm, yshift=0.55cm}]
      \addplot+[mark = *, mark size=1pt] table[x=num_regions, y=CO, col sep=comma] {figures/sp_metrics/VOC_Context_ea.csv};
      \addplot+[mark = *, mark size=1pt] table[x=num_regions, y=CO, col sep=comma] {figures/sp_metrics/VOC_Context_slic.csv};
      \addplot+[mark = *, mark size=1pt] table[x=num_regions, y=CO, col sep=comma] {figures/sp_metrics/VOC_Context_snic.csv};
      \addplot+[mark = *, mark size=1pt] table[x=num_regions, y=CO, col sep=comma] {figures/sp_metrics/VOC_Context_lns.csv};
      \addplot+[mark = *, mark size=1pt, color=cyan, mark options={fill=cyan}] table[x=num_regions, y=CO, col sep=comma] {figures/sp_metrics/VOC_Context_ddseg.csv};
    \hfill
    \end{groupplot}
  \end{tikzpicture}
\caption{Quantitative superpixel quality evaluation against the number of superpixels (\#SP) on the test set of the BSDS500 and the PASCAL-Context.}
\label{fig:natural_quant}
\end{figure*}

\subsubsection{Natural Image Results}
\begin{figure}
  \centering
  \begin{subfigure}{0.49\linewidth}
    \includegraphics[width=1.0\linewidth]{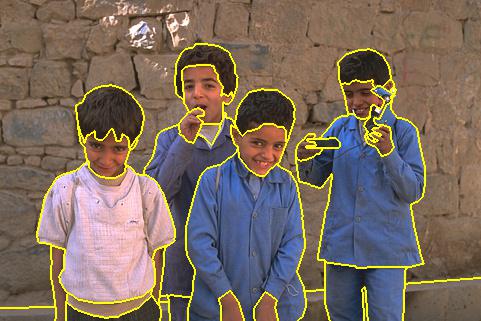}
    \caption{Sample Ground Truth}
    \label{fig:qual-bsds-e}
  \end{subfigure}
  \hfill
    \begin{subfigure}{0.49\linewidth}
    \includegraphics[width=1.0\linewidth]{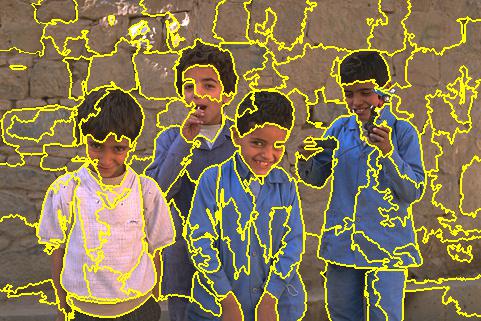}
    \caption{SNIC}
    \label{fig:qual-bsds-d}
  \end{subfigure}
  \hfill
    \begin{subfigure}{0.49\linewidth}
    \includegraphics[width=1.0\linewidth]{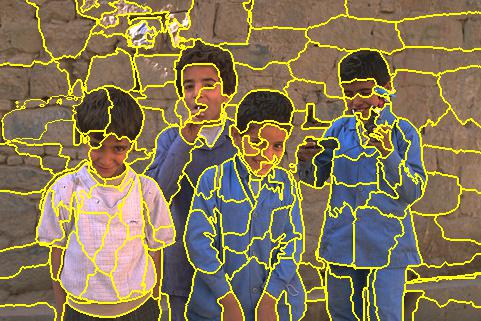}
    \caption{EA}
    \label{fig:qual-bsds-f}
  \end{subfigure}
  \hfill
  \begin{subfigure}{0.49\linewidth}
    \includegraphics[width=1.0\linewidth]{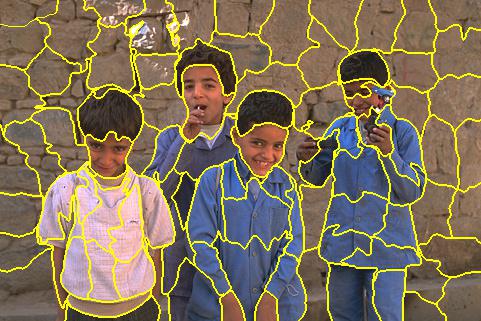}
    \caption{Ours}
    \label{fig:qual-bsds-b}
  \end{subfigure}
\caption{Qualitative Comparison of Superpixel Algorithms on an image from BSDS500.}
\label{fig:qual_bsds}
\end{figure}

\begin{figure}
  \centering
  \begin{subfigure}{0.48\linewidth}
    \includegraphics[width=1.0\linewidth]{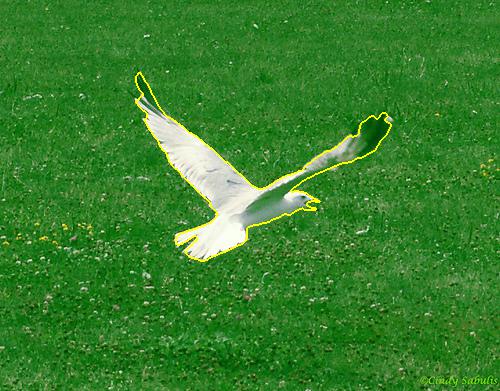}
    \caption{Ground Truth}
    \label{fig:qual-voc-e}
  \end{subfigure}
  \hfill
  \begin{subfigure}{0.48\linewidth}
    \includegraphics[width=1.0\linewidth]{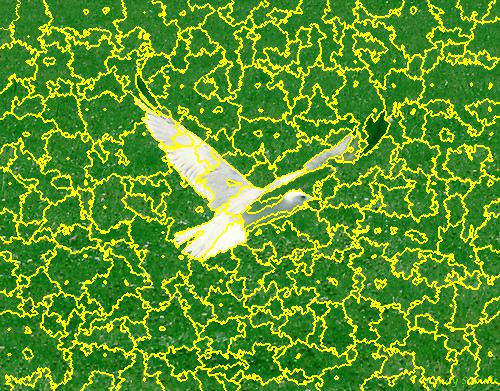}
    \caption{SNIC}
    \label{fig:qual-voc-d}
  \end{subfigure}
  \hfill
    \begin{subfigure}{0.48\linewidth}
    \includegraphics[width=1.0\linewidth]{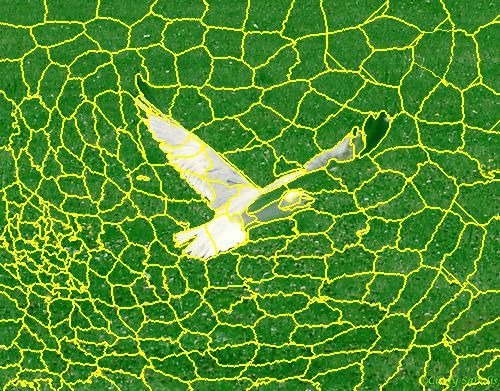}
    \caption{EA}
    \label{fig:qual-voc-f}
  \end{subfigure}
  \hfill
  \begin{subfigure}{0.48\linewidth}
    \includegraphics[width=1.0\linewidth]{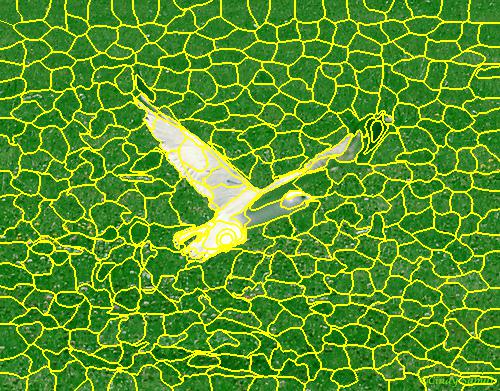}
    \caption{Ours}
    \label{fig:qual-voc-b}
  \end{subfigure}
  \hfill
\caption{Qualitative Comparison of Superpixel Algorithms on an image from PASCAL-Context.}
\label{fig:qual_voc}
\end{figure}

As we can see in \Cref{fig:natural_quant}, our approach consistently outperforms the benchmark methods in terms of USE and ASA. It shows the best edge adherence in terms of BR on the PASCAL-Context dataset and second-best on the BSDS500, where SNIC shows better edge behaviour however at the cost of a factor of two drop in compactness compared to our method. In terms of compactness our approach closely matches the performance of the edge-aware untrained net EA but outperforms it on other metrics. In \Cref{fig:qual_bsds} superpixel sets with a cluster parameter of 100 on an image of the BSDS500 can be compared. We can observe that our approach favour building superpixels around globally significant edges. That is why even at a relatively low superpixel cluster parameter our method produces visually pleasing superpixels with smooth boundaries and high similarity to the provided ground truth. It is also visualized how at high variance image regions other methods implicitly generate more superpixels than desired by not enforcing connectivity. In \Cref{fig:qual_voc} superpixel partitions of an image from the PASCAL-Context dataset are displayed. This example shows how other methods focus on arbitrary edges in the background of an image that are irrelevant with respect to the ground truth segmentation. Our method discards arbitrary edges in noisy backgrounds in exchange for more compactness.


\begin{figure*}[h]
  \centering
  \begin{subfigure}{0.19\linewidth}
    \includegraphics[width=1.0\linewidth]{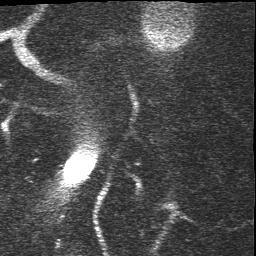}
    \caption{Original Image}
    \label{fig:qual-deepvess-a}
  \end{subfigure}
  \hfill
  \begin{subfigure}{0.19\linewidth}
    \includegraphics[width=1.0\linewidth]{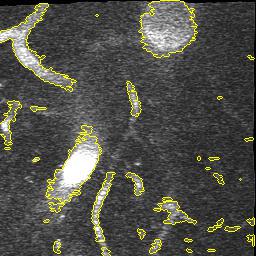}
    \caption{SNIC, DICE 81.6\%}
    \label{fig:qual-deepvess-c}
  \end{subfigure}
  \hfill
  \begin{subfigure}{0.19\linewidth}
    \includegraphics[width=1.0\linewidth]{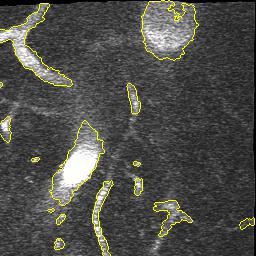}
    \caption{EA, DICE 81.3\%}
    \label{fig:qual-deepvess--b}
  \end{subfigure}
  \hfill
  \begin{subfigure}{0.19\linewidth}
    \includegraphics[width=1.0\linewidth]{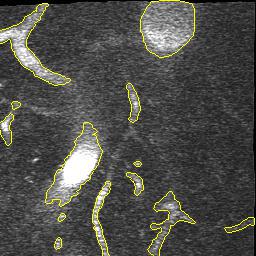}
    \caption{Ours, DICE 87.9\%}
    \label{fig:qual-deepvess--a}
  \end{subfigure}
  \hfill
  \begin{subfigure}{0.19\linewidth}
    \includegraphics[width=1.0\linewidth]{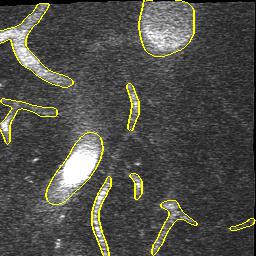}
    \caption{Ground Truth}
    \label{fig:qual-deepvess-d}
  \end{subfigure}
  \hfill
\caption{Qualitative Comparison of Superpixel Algorithms on an image from DeepVess.}
\label{fig:qual_deepvess}
\end{figure*}

\subsubsection{Microscopy Foreground Segmentation}

The results in \Cref{tab:deepvess} show that our approach performs best across other superpixel algorithms in an end-to-end segmentation pipeline. Based on this observation we conclude that the slight drop in compactness we observed in the natural image experiments with respect to EA does not affect the quality of post-processing. Moreover, the results stress our method's robustness with respect to noise. It is also worth noting that it closely matches the performance of a trained neural networkand the second human annotation on this dataset. While the Double-DIP seems to work on natural images with a saliency estimate, it does not converge on noisy microscopic data which is why we decided to stop the optimization after 500 steps. In \Cref{fig:qual_deepvess} we can see superpixel based foreground predictions generated from a DeepVess slice. While SNIC fails to match superpixels to vessel structures and overfits to noise artifacts, EA and our method produce smoother superpixels. However, the Deep Decoder is better at detecting low-intensity shapes and adhering to vessel boundaries, as it can be seen around the ground truth objects in the top left part in \Cref{fig:qual_deepvess}, which is why the segmentation pipeline works best with our method as a superpixel generator.

\begin{table}
  \centering
  \begin{tabular}{@{}lcccc@{}}
    \toprule
    Method & $S_e$ & $S_p$ & $DC$ & $JI$ \\
        \midrule
                       
                            Li &         67.5 &         \textbf{98.0} &  72.3 &     57.2 \\
                     BM3D + Li &         \textbf{88.3} &         97.5 &  79.8 &     66.7 \\
               Li + Double-DIP &         78.2 &         97.7 &  76.8 &     63.0 \\
               EA + Li &         82.3 &         97.7 &  79.2 &     65.7 \\
                     SNIC + Li &         80.5 &         97.7 &  77.9 &     63.9 \\
                    Ours + Li &         85.1 &         97.9 &  \textbf{81.6} &     \textbf{69.1} \\
        \midrule
        Supervised CNN &         90.0 &         97.0 &  81.6 &     69.2 \\
        
        Second Human Annotator &         81.1 &         98.7 &  82.4 &     70.4 \\
        \bottomrule
  \end{tabular}
  \caption{Quantitative results on the DeepVess dataset in \%.}
  \label{tab:deepvess}
\end{table}

\subsection{Ablation Study}
\label{sec:ablation}
We conducted an ablation study with respect to the compactness parameter $\lossweight$, the number of deep decoders $\ddnum$ and the blur factor $\blurfactor$ on 25 images of the BSDS500 train set. Quantitative results for a few relevant hyperparameter combinations at a cluster parameter of 400 can be found in \Cref{tab:ablation}. More results are included in the supplementary material. 
We experimentally found that decreasing the compactness parameter $\lossweight$ improves the compactness at the expense of worse boundary recall. This was expected as a low compactness parameter $\lossweight$ incentivises the deep decoder to encode more edge than spatial information into its last layer. A growing blur factor $\blurfactor$ makes the deep decoder neglect more edges in its reconstruction and hence, increases compactness and under segmentation error but reduces boundary adherence. An increase in the number of deep decoders from $\ddnum=1$ to $\ddnum=3$, significantly increases compactness because for each deep decoder we draw a random offset $\offset$ for the sinusoidials in the positional encoding. Other metrics only change marginally with a growing number of deep decoders. We chose a hyperparameter combination where no pareto improvement can be found with respect to the metrics and execution time.

\begin{table}
  \centering
  \begin{tabular}{@{}lcc|cccc@{}}
    \toprule
    $\lossweight$ & $\blurfactor$ & $\ddnum$  &      USE &       BR &      ASA &       CO \\
        \midrule
        \textit{0.05} \tikzmark{numiter3}& 0.0001 &       3 &   0.37 & 0.84 \tikzmark{numiterbr3} & 0.96 & 0.45 \tikzmark{numiterco3}\\
        \textit{0.10} \tikzmark{numiter2}& 0.0001 &       3 &  0.37 & 0.83 \tikzmark{numiterbr2}& 0.96 & 0.52 \tikzmark{numiterco2}\\
        \textit{0.15} \tikzmark{numiter1} & 0.0001 &       3 &       0.37 & 0.80 \tikzmark{numiterbr1} & 0.96 & 0.56 \tikzmark{numiterco1}\\
        \midrule
         0.9 & \textit{0.00005} \tikzmark{blur1}&       3 &   0.38 \tikzmark{bluruse1}& 0.84 \tikzmark{blurbr1}& 0.96 & 0.46 \tikzmark{blurco1}\\
         0.9 & \textit{0.00010} \tikzmark{blur2}&       3 &  0.37 \tikzmark{bluruse2}& 0.83 \tikzmark{blurbr2}& 0.96 & 0.52 \tikzmark{blurco2}\\
         0.9 & \textit{0.00020} \tikzmark{blur3}&       3 &    0.35 \tikzmark{bluruse3}& 0.80 \tikzmark{blurbr3}& 0.96 & 0.56 \tikzmark{blurco3}\\
         \midrule
         0.9 & 0.0001 &       \textit{1} \tikzmark{numdd1}&   0.39 \tikzmark{numdduse1}& 0.82 & 0.96 & 0.47 \tikzmark{numddco1}\\
         0.9 & 0.0001 &       \textit{3} \tikzmark{numdd2}&  0.37 \tikzmark{numdduse2}& 0.83 & 0.96 & 0.52 \tikzmark{numddco2}\\
         0.9 & 0.0001 &       \textit{5} \tikzmark{numdd3}&  0.36 \tikzmark{numdduse3}& 0.82 & 0.97 & 0.53 \tikzmark{numddco3}\\
        \bottomrule

        \begin{tikzpicture}[overlay,remember picture]
        \draw[-Triangle, color=black] ($(pic cs:numiterbr1)+(2pt,0.5ex)$) to ($(pic cs:numiterbr2)+(2pt,0.5ex)$) to ($(pic cs:numiterbr3)+(2pt,1ex)$);
        \draw[-Triangle, color=black] ($(pic cs:numiterco3)+(2pt,1ex)$) to ($(pic cs:numiterco2)+(2pt,0.5ex)$) to ($(pic cs:numiterco1)+(2pt,0.5ex)$);
        
        \draw[-Triangle, color=black] ($(pic cs:bluruse3)+(2pt,0.5ex)$) to ($(pic cs:bluruse2)+(2pt,0.5ex)$) to ($(pic cs:bluruse1)+(2pt,1ex)$);
        \draw[-Triangle, color=black] ($(pic cs:blurbr3)+(2pt,0.5ex)$) to ($(pic cs:blurbr2)+(2pt,0.5ex)$) to ($(pic cs:blurbr1)+(2pt,1ex)$);
        \draw[-Triangle, color=black] ($(pic cs:blurco1)+(2pt,1ex)$) to ($(pic cs:blurco2)+(2pt,0.5ex)$) to ($(pic cs:blurco3)+(2pt,0.5ex)$);
        
        \draw[-Triangle, color=black] ($(pic cs:numdduse3)+(2pt,0.5ex)$) to ($(pic cs:numdduse2)+(2pt,0.5ex)$) to ($(pic cs:numdduse1)+(2pt,1ex)$);
        \draw[-Triangle, color=black] ($(pic cs:numdduse3)+(2pt,0.5ex)$) to ($(pic cs:numdduse2)+(2pt,0.5ex)$) to ($(pic cs:numdduse1)+(2pt,1ex)$);
        \draw[-Triangle, color=black] ($(pic cs:numddco1)+(2pt,1ex)$) to ($(pic cs:numddco2)+(2pt,0.5ex)$) to ($(pic cs:numddco3)+(2pt,0.5ex)$);
        
        
        \end{tikzpicture}
  \end{tabular}
  \caption{\textbf{Selected Ablation Results} on 25 images of the BSDS500 train set for 400 clusters.}
  \label{tab:ablation}
\end{table}

\section{Conclusion}
We exploit a non-convolutional deep image prior's ability to create edge sparse image reconstructions to extract pixel embeddings from its last hidden layer. We further encode spatial information in these activation maps by designing a positional awareness loss based on a sinusoidal positional encoding. From these pixel embeddings we generate high-quality superpixels via a SLIC-like clustering step. The resulting superpixels are more likely to focus on global and significant image edges which are more relevant for further post-processing such as segmentation. Our proposed method shows state-of-the-art performance on two natural image datasets and on one microscopic dataset as part of a downstream segmentation task.
{\small
\bibliographystyle{ieee_fullname}
\bibliography{egbib}
}

\end{document}


\title{Supplementary: Unsupervised Superpixel Generation using Edge-Sparse Embedding}

\author{Jakob Geusen\textsuperscript{1}, Gustav Bredell\textsuperscript{2}, Tianfei Zhou\textsuperscript{2}, Ender Konukoglu\textsuperscript{2}\\
Department of Information Technology and Electrical Engineering\\
ETH-Zurich, Zurich, Switzerland\\
\textsuperscript{1}{\tt\small jgeusen@student.ethz.ch}\\
\textsuperscript{2}{\tt\small \{gustav.bredell, tianfei.zhou, ender.konukoglu\}@vision.ee.ethz.ch}\\
}
\maketitle
\appendix

\section{DeepVess Decoder Architecture Modification}
\label{sec:deepvess}

By noticing that the DeepVess slices are very noisy and depict smooth vessel structures, we can further improve our method by downsizing our deep decoder base architecture and increasing the input blur. We rely on the Deep Decoder's denoising properties \cite{heckel_deep_2019} to improve our segmentation performance with respect to point-wise artifacts. By decreasing the decoder's number of layers and channels we further decrease the number of edges and the amount of detail in the decoder's reconstruction and its last layer. In combination with an increased input blur we can also observe smoother boundaries in the superpixel partitions and hence, in the foreground segmentation. We reduce the number of channels from $\numchannels=128$ to $\numchannels=32$, the number of blocks consisting of linear channel combinations, up-sampling, Re-LU and batch normalization from 5 to 4 and increase the blur factor from $\blurfactor=0.0001$ to $\blurfactor=0.0002$. We also increase the number of deep decoders from $\ddnum=3$ to $\ddnum=5$, because we notice an increased variance in the feature generation with smaller decoders. This however, affects the run time only marginally as it takes less time to fit a downsized decoder.

\begin{figure*}[h]
  \centering
  \begin{subfigure}{0.16\linewidth}
    \includegraphics[width=1.0\linewidth]{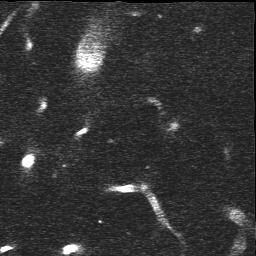}
    \caption{Original Image}
    \label{fig:qual-dv-26-a}
  \end{subfigure}
  \hfill
    \begin{subfigure}{0.16\linewidth}
    \includegraphics[width=1.0\linewidth]{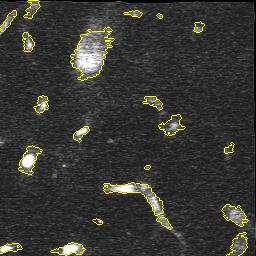}
    \caption{SNIC}
    \label{fig:qual-dv-26-b}
  \end{subfigure}
  \hfill
    \begin{subfigure}{0.16\linewidth}
    \includegraphics[width=1.0\linewidth]{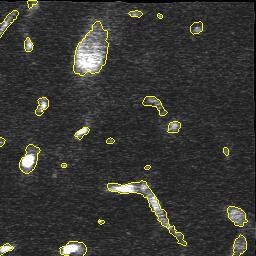}
    \caption{EA}
    \label{fig:qual-dv-26-c}
  \end{subfigure}
  \hfill
  \begin{subfigure}{0.16\linewidth}
    \includegraphics[width=1.0\linewidth]{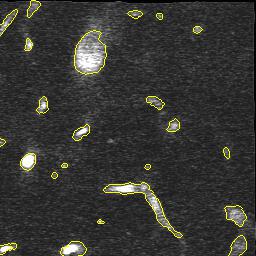}
    \caption{Ours}
    \label{fig:qual-dv-26-d}
  \end{subfigure}
    \begin{subfigure}{0.16\linewidth}
    \includegraphics[width=1.0\linewidth]{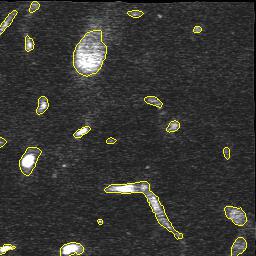}
    \caption{Ours (Downsized)}
    \label{fig:qual-dv-26-e}
  \end{subfigure}
    \begin{subfigure}{0.16\linewidth}
    \includegraphics[width=1.0\linewidth]{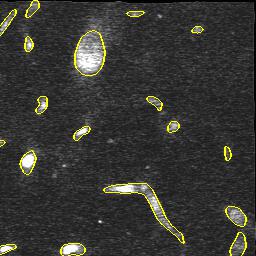}
    \caption{Ground Truth}
    \label{fig:qual-dv-26-f}
  \end{subfigure}
  
  \begin{subfigure}{0.16\linewidth}
    \includegraphics[width=1.0\linewidth]{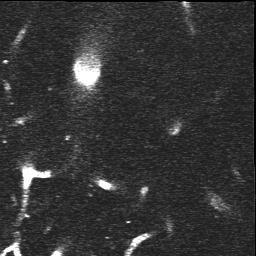}
    \caption{Original Image}
    \label{fig:qual-dv-42-a}
  \end{subfigure}
  \hfill
    \begin{subfigure}{0.16\linewidth}
    \includegraphics[width=1.0\linewidth]{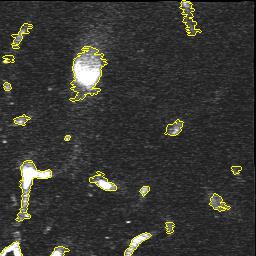}
    \caption{SNIC}
    \label{fig:qual-dv-42-b}
  \end{subfigure}
  \hfill
    \begin{subfigure}{0.16\linewidth}
    \includegraphics[width=1.0\linewidth]{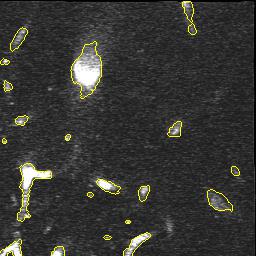}
    \caption{EA}
    \label{fig:qual-dv-42-c}
  \end{subfigure}
  \hfill
  \begin{subfigure}{0.16\linewidth}
    \includegraphics[width=1.0\linewidth]{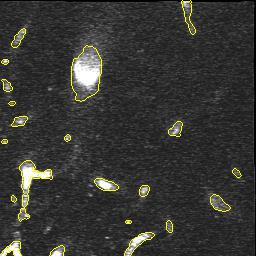}
    \caption{Ours}
    \label{fig:qual-dv-42-d}
  \end{subfigure}
    \begin{subfigure}{0.16\linewidth}
    \includegraphics[width=1.0\linewidth]{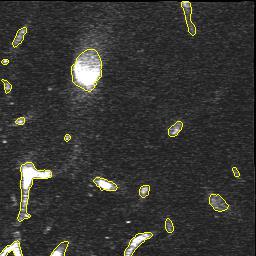}
    \caption{Ours (Downsized)}
    \label{fig:qual-dv-42-e}
  \end{subfigure}
    \begin{subfigure}{0.16\linewidth}
    \includegraphics[width=1.0\linewidth]{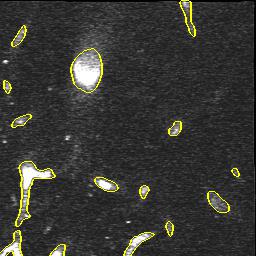}
    \caption{Ground Truth}
    \label{ffig:qual-dv-42-af}
  \end{subfigure}

  \begin{subfigure}{0.16\linewidth}
    \includegraphics[width=1.0\linewidth]{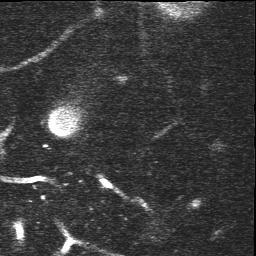}
    \caption{Original Image}
    \label{fig:qual-dv-102-a}
  \end{subfigure}
  \hfill
    \begin{subfigure}{0.16\linewidth}
    \includegraphics[width=1.0\linewidth]{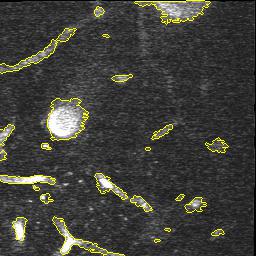}
    \caption{SNIC}
    \label{fig:qual-dv-102-b}
  \end{subfigure}
  \hfill
    \begin{subfigure}{0.16\linewidth}
    \includegraphics[width=1.0\linewidth]{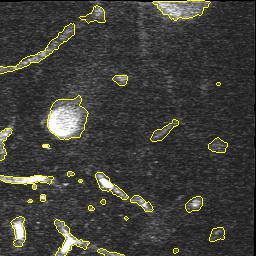}
    \caption{EA}
    \label{fig:qual-dv-102-c}
  \end{subfigure}
  \hfill
  \begin{subfigure}{0.16\linewidth}
    \includegraphics[width=1.0\linewidth]{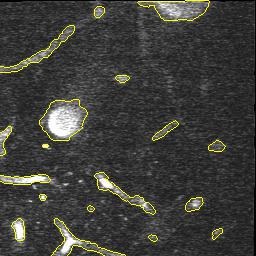}
    \caption{Ours}
    \label{fig:qual-dv-102-d}
  \end{subfigure}
    \begin{subfigure}{0.16\linewidth}
    \includegraphics[width=1.0\linewidth]{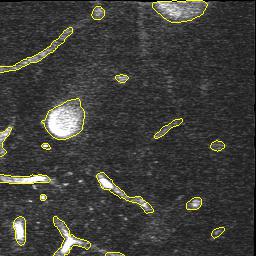}
    \caption{Ours (Downsized)}
    \label{fig:qual-dv-102-e}
  \end{subfigure}
    \begin{subfigure}{0.16\linewidth}
    \includegraphics[width=1.0\linewidth]{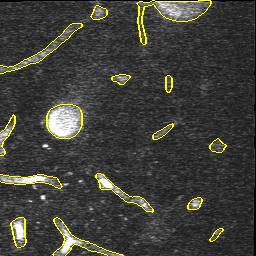}
    \caption{Ground Truth}
    \label{fig:qual-dv-102-f}
  \end{subfigure}
  
\caption{Qualitative Comparison of Superpixel Algorithms incorporated in a foreground segementation pipeline on the DeepVess dataset.}
\label{fig:qual_deepvess}
\end{figure*}

In \Cref{fig:qual-dv-26-d} and \Cref{fig:qual-dv-26-e} we can observe, how the smaller decoder architecture causes some of the artifacts not to get get classified as foreground. Smoother boundaries also result in better edge adherence around large, high-intensity blobs in comparison to the original method, as seen in \Cref{fig:qual-dv-42-e} and \Cref{fig:qual-dv-102-e}. This also leads to better $S_e$, $DC$ and $JI$, with only a minor decline in $S_p$, as displayed in \Cref{tab:deepvess}.

\begin{table}[h]
  \centering
  \begin{tabular}{@{}lcccc@{}}
    \toprule
    Method & $S_e$ & $S_p$ & $DC$ & $JI$ \\
        \midrule
                       
                            Li &         67.5 &         \textbf{98.0} &  72.3 &     57.2 \\
                     BM3D + Li &         88.3 &         97.5 &  79.8 &     66.7 \\
               Li + Double-DIP &         78.2 &         97.7 &  76.8 &     63.0 \\
               EA + Li &         82.3 &         97.7 &  79.2 &     65.7 \\
                     SNIC + Li &         80.5 &         97.7 &  77.9 &     63.9 \\
                    Ours + Li &         85.1 &         97.9 &  81.6 &     69.1 \\
                    Ours (Downsized) + Li &         \textbf{88.4} &         97.8 &  \textbf{82.4} &     \textbf{70.1} \\
        \midrule
        Supervised CNN &         90.0 &         97.0 &  81.6 &     69.2 \\
        
        Second Human Annotator &         81.1 &         98.7 &  82.4 &     70.4 \\
        \bottomrule
  \end{tabular}
  \caption{Quantitative results on the DeepVess dataset in \%.}
  \label{tab:deepvess}
\end{table}

\clearpage
\section{Additional Ablation Results}

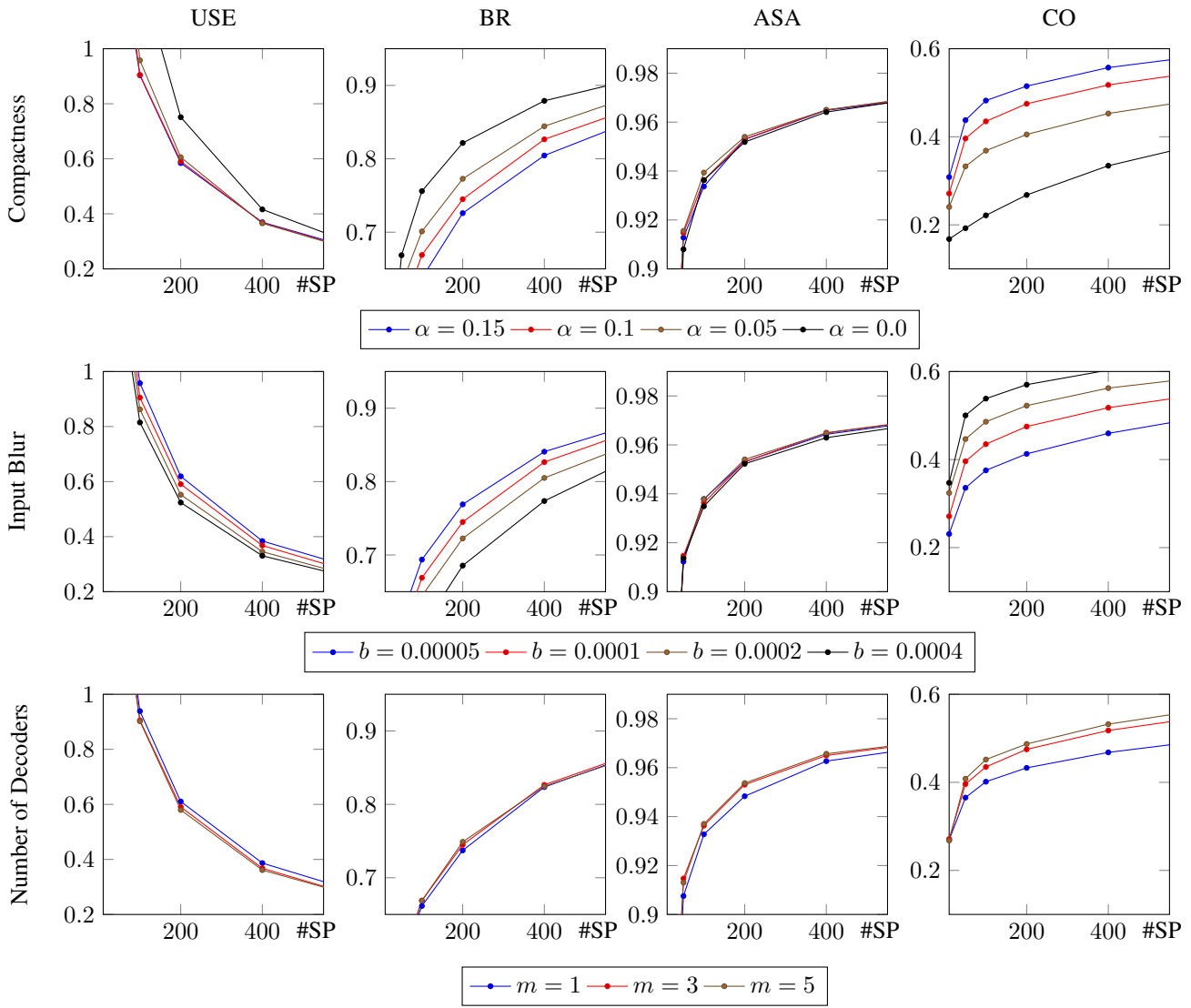
\begin{figure*}[h]
    \centering
  \begin{tikzpicture}
    \begin{groupplot}[
      group style={group size=4 by 3, vertical sep=1.5cm, horizontal sep = 0.9cm},
      width=4.8cm, height=4.8cm
    ]
    \nextgroupplot[title=USE, ymin = 0.2, ymax= 1.0, xmin=10, xmax=550, ylabel=Compactness, xlabel={\#SP},
    xlabel style={xshift=1.5cm, yshift=0.55cm}]
      \addplot+[mark = *, mark size=1pt] table[x=num_clusters, y=USE, col sep=comma] {figures/ablation_plots.csv};
      \addplot+[mark = *, mark size=1pt] table[x=num_clusters, y=USE_lw0.9_bl0.0001_ndd3, col sep=comma] {figures/ablation_plots.csv};
      \addplot+[mark = *, mark size=1pt] table[x=num_clusters, y=USE_lw0.95_bl0.0001_ndd3, col sep=comma] {figures/ablation_plots.csv};
      \addplot+[mark = *, mark size=1pt] table[x=num_clusters, y=USE_lw1.0_bl0.0001_ndd3, col sep=comma] {figures/ablation_plots.csv};
    \hfill
    \nextgroupplot[title=BR, ymin = 0.65, ymax= 0.95, xmin=10, xmax=550, xlabel={\#SP},
    xlabel style={xshift=1.5cm, yshift=0.55cm}]
      \addplot+[mark = *, mark size=1pt] table[x=num_clusters, y=BR, col sep=comma] {figures/ablation_plots.csv};
      \addplot+[mark = *, mark size=1pt] table[x=num_clusters, y=BR_lw0.9_bl0.0001_ndd3, col sep=comma] {figures/ablation_plots.csv};
      \addplot+[mark = *, mark size=1pt] table[x=num_clusters, y=BR_lw0.95_bl0.0001_ndd3, col sep=comma] {figures/ablation_plots.csv};
      \addplot+[mark = *, mark size=1pt] table[x=num_clusters, y=BR_lw1.0_bl0.0001_ndd3, col sep=comma] {figures/ablation_plots.csv};
    \hfill
    \nextgroupplot[title=ASA, ymin = 0.9, ymax= 0.99, xmin=10, xmax=550, xlabel={\#SP},
    xlabel style={xshift=1.5cm, yshift=0.55cm}]
      \addplot+[mark = *, mark size=1pt] table[x=num_clusters, y=ASA, col sep=comma] {figures/ablation_plots.csv};
      \addplot+[mark = *, mark size=1pt] table[x=num_clusters, y=ASA_lw0.9_bl0.0001_ndd3, col sep=comma] {figures/ablation_plots.csv};
      \addplot+[mark = *, mark size=1pt] table[x=num_clusters, y=ASA_lw0.95_bl0.0001_ndd3, col sep=comma] {figures/ablation_plots.csv};
      \addplot+[mark = *, mark size=1pt] table[x=num_clusters, y=ASA_lw1.0_bl0.0001_ndd3, col sep=comma] {figures/ablation_plots.csv};
    \hfill
    \nextgroupplot[title=CO, ymin = 0.1, ymax= 0.6, xmin=10, xmax=550, xlabel={\#SP},
    xlabel style={xshift=1.5cm, yshift=0.55cm}, legend style={legend columns=4,at={(-8.6cm,-0.6cm)},anchor=north west}, xlabel={\#SP},
    xlabel style={xshift=0cm, yshift=0cm}]
      \addplot+[mark = *, mark size=1pt] table[x=num_clusters, y=CO, col sep=comma] {figures/ablation_plots.csv};
      \addplot+[mark = *, mark size=1pt] table[x=num_clusters, y=CO_lw0.9_bl0.0001_ndd3, col sep=comma] {figures/ablation_plots.csv};
      \addplot+[mark = *, mark size=1pt] table[x=num_clusters, y=CO_lw0.95_bl0.0001_ndd3, col sep=comma] {figures/ablation_plots.csv};
      \addplot+[mark = *, mark size=1pt] table[x=num_clusters, y=CO_lw1.0_bl0.0001_ndd3, col sep=comma] {figures/ablation_plots.csv};
    \hfill
      \legend{$\lossweight=0.15$, $\lossweight=0.1$, $\lossweight=0.05$, $\lossweight=0.0$}

    \nextgroupplot[ymin = 0.2, ymax= 1.0, xmin=10, xmax=550, ylabel=Input Blur, xlabel={\#SP},
    xlabel style={xshift=1.5cm, yshift=0.55cm}]
      \addplot+[mark = *, mark size=1pt] table[x=num_clusters, y=USE_lw0.9_bl5e-05_ndd3, col sep=comma] {figures/ablation_plots.csv};
      \addplot+[mark = *, mark size=1pt] table[x=num_clusters, y=USE_lw0.9_bl0.0001_ndd3, col sep=comma] {figures/ablation_plots.csv};
      \addplot+[mark = *, mark size=1pt] table[x=num_clusters, y=USE_lw0.9_bl0.0002_ndd3, col sep=comma] {figures/ablation_plots.csv};
      \addplot+[mark = *, mark size=1pt] table[x=num_clusters, y=USE_lw0.9_bl0.0004_ndd3, col sep=comma] {figures/ablation_plots.csv};
    \hfill
    \nextgroupplot[ymin = 0.65, ymax= 0.95, xmin=10, xmax=550, xlabel={\#SP},
    xlabel style={xshift=1.5cm, yshift=0.55cm}]
      \addplot+[mark = *, mark size=1pt] table[x=num_clusters, y=BR_lw0.9_bl5e-05_ndd3, col sep=comma] {figures/ablation_plots.csv};
      \addplot+[mark = *, mark size=1pt] table[x=num_clusters, y=BR_lw0.9_bl0.0001_ndd3, col sep=comma] {figures/ablation_plots.csv};
      \addplot+[mark = *, mark size=1pt] table[x=num_clusters, y=BR_lw0.9_bl0.0002_ndd3, col sep=comma] {figures/ablation_plots.csv};
      \addplot+[mark = *, mark size=1pt] table[x=num_clusters, y=BR_lw0.9_bl0.0004_ndd3, col sep=comma] {figures/ablation_plots.csv};
    \hfill
    \nextgroupplot[ymin = 0.9, ymax= 0.99, xmin=10, xmax=550, legend style={legend columns=4,at={(-5.3cm, 3.35cm)},anchor=north west}, xlabel={\#SP},
    xlabel style={xshift=1.5cm, yshift=0.55cm}]
      \addplot+[mark = *, mark size=1pt] table[x=num_clusters, y=ASA_lw0.9_bl5e-05_ndd3, col sep=comma] {figures/ablation_plots.csv};
      \addplot+[mark = *, mark size=1pt] table[x=num_clusters, y=ASA_lw0.9_bl0.0001_ndd3, col sep=comma] {figures/ablation_plots.csv};
      \addplot+[mark = *, mark size=1pt] table[x=num_clusters, y=ASA_lw0.9_bl0.0002_ndd3, col sep=comma] {figures/ablation_plots.csv};
      \addplot+[mark = *, mark size=1pt] table[x=num_clusters, y=ASA_lw0.9_bl0.0004_ndd3, col sep=comma] {figures/ablation_plots.csv};
      \legend{$\blurfactor=0.00005$, $\blurfactor=0.0001$, $\blurfactor=0.0002$, $\blurfactor=0.0004$}
    \hfill
    \nextgroupplot[ymin = 0.1, ymax= 0.6, xmin=10, xmax=550, xlabel={\#SP},
    xlabel style={xshift=1.5cm, yshift=0.55cm}]
      \addplot+[mark = *, mark size=1pt] table[x=num_clusters, y=CO_lw0.9_bl5e-05_ndd3, col sep=comma] {figures/ablation_plots.csv};
      \addplot+[mark = *, mark size=1pt] table[x=num_clusters, y=CO_lw0.9_bl0.0001_ndd3, col sep=comma] {figures/ablation_plots.csv};
      \addplot+[mark = *, mark size=1pt] table[x=num_clusters, y=CO_lw0.9_bl0.0002_ndd3, col sep=comma] {figures/ablation_plots.csv};
      \addplot+[mark = *, mark size=1pt] table[x=num_clusters, y=CO_lw0.9_bl0.0004_ndd3, col sep=comma] {figures/ablation_plots.csv};
    \hfill

    \nextgroupplot[ymin = 0.2, ymax= 1.0, xmin=10, xmax=550, ylabel=Number of Decoders, xlabel={\#SP},
    xlabel style={xshift=1.5cm, yshift=0.55cm}]
      \addplot+[mark = *, mark size=1pt] table[x=num_clusters, y=USE_lw0.9_bl0.0001_ndd1, col sep=comma] {figures/ablation_plots.csv};
      \addplot+[mark = *, mark size=1pt] table[x=num_clusters, y=USE_lw0.9_bl0.0001_ndd3, col sep=comma] {figures/ablation_plots.csv};
      \addplot+[mark = *, mark size=1pt] table[x=num_clusters, y=USE_lw0.9_bl0.0001_ndd5, col sep=comma] {figures/ablation_plots.csv};
    \hfill
    \nextgroupplot[ymin = 0.65, ymax= 0.95, xmin=10, xmax=550, xlabel={\#SP},
    xlabel style={xshift=1.5cm, yshift=0.55cm}]
      \addplot+[mark = *, mark size=1pt] table[x=num_clusters, y=BR_lw0.9_bl0.0001_ndd1, col sep=comma] {figures/ablation_plots.csv};
      \addplot+[mark = *, mark size=1pt] table[x=num_clusters, y=BR_lw0.9_bl0.0001_ndd3, col sep=comma] {figures/ablation_plots.csv};
      \addplot+[mark = *, mark size=1pt] table[x=num_clusters, y=BR_lw0.9_bl0.0001_ndd5, col sep=comma] {figures/ablation_plots.csv};
    \hfill
    \nextgroupplot[ymin = 0.9, ymax= 0.99, xmin=10, xmax=550, legend style={legend columns=3, at={(-3cm, 3.35cm)},anchor=north west}, xlabel={\#SP},
    xlabel style={xshift=1.5cm, yshift=0.55cm}]
      \addplot+[mark = *, mark size=1pt] table[x=num_clusters, y=ASA_lw0.9_bl0.0001_ndd1, col sep=comma] {figures/ablation_plots.csv};
      \addplot+[mark = *, mark size=1pt] table[x=num_clusters, y=ASA_lw0.9_bl0.0001_ndd3, col sep=comma] {figures/ablation_plots.csv};
      \addplot+[mark = *, mark size=1pt] table[x=num_clusters, y=ASA_lw0.9_bl0.0001_ndd5, col sep=comma] {figures/ablation_plots.csv};
      \legend{$\ddnum=1$, $\ddnum=3$, $\ddnum=5$}
    \hfill
    \nextgroupplot[ymin = 0.1, ymax= 0.6, xmin=10, xmax=550, xlabel={\#SP},
    xlabel style={xshift=1.5cm, yshift=0.55cm}]
      \addplot+[mark = *, mark size=1pt] table[x=num_clusters, y=CO_lw0.9_bl0.0001_ndd1, col sep=comma] {figures/ablation_plots.csv};
      \addplot+[mark = *, mark size=1pt] table[x=num_clusters, y=CO_lw0.9_bl0.0001_ndd3, col sep=comma] {figures/ablation_plots.csv};
      \addplot+[mark = *, mark size=1pt] table[x=num_clusters, y=CO_lw0.9_bl0.0001_ndd5, col sep=comma] {figures/ablation_plots.csv};
    \hfill

    \end{groupplot}
  \end{tikzpicture}
\caption{Quantitative superpixel quality evaluation against the number of superpixels (\#SP) on 25 images of the BSDS500 train set.}
\label{fig:ablation}
\end{figure*}

In \Cref{fig:ablation} we can see how the superpixel quality metrics change with respect to alterations of the compactness parameter $\lossweight$, the input blur factor $\blurfactor$ and the number $\ddnum$ of deep decoders incorporated in our method. Increasing the compactness parameter enforces more spatial information and less edge information in the deep decoder's last layer and therefore, we can exchange more compactness for less edge adherence. A notable difference in CO and BR between $\lossweight=0$ and $\lossweight=0.051$ can be spotted in \Cref{fig:ablation} because a higher superpixel compactness can help decreasing USE as we allow less boundary leakage by restricting the superpixels spatially. In \Cref{fig:ablation} we can also examine, how increasing the input blur makes the deep decoder neglect more edges and consequently, increases the CO and USE of the produced superpixels. Regarding the number of deep decoders, we can conclude that the quality metrics quickly converge. While the jump in superpixel quality from using $\ddnum=1$ to $\ddnum=3$ deep decoders can be considered significant, especially, regarding compactness, we argue that further increasing the number of deep decoders (and the method's execution time) cannot be justified. This statement holds for larger Deep Decoder architectures that we employ for natural images with a high signal-to-noise ratio, \eg BSDS500 and PASCAL-Context, because the reconstructions vary not as much from network to network in comparison to highly under-parameterized Deep Decoders fitted to noisy images (\cref{sec:deepvess}).

\clearpage
\section{Further Qualitative Results}
\begin{figure*}[h]
    \centering
  \begin{tikzpicture}
    \begin{groupplot}[
      group style={group size=1 by 1, vertical sep=1.5cm, horizontal sep = 0.9cm},
      width=6cm, height=6cm
    ]
    \nextgroupplot[title=USE, xmin=0, xmax=350, ymin=0, ymax=350, ylabel=Number of Superpixels, xlabel={Cluster Number Input}, domain=0:600, legend style={legend columns=2,at={(5.5cm,3cm)},anchor=north west}]
      \addplot+[mark = *, mark size=1pt] table[x=num_clusters, y=num_regions_ea, col sep=comma] {figures/supplementary/cluster_region_plot.csv};
      \addplot+[mark = *, mark size=1pt] table[x=num_clusters, y=num_regions, col sep=comma] {figures/supplementary/cluster_region_plot.csv};
      \addplot+[mark = *, mark size=1pt] table[x=num_clusters, y=num_regions_snic, col sep=comma] {figures/supplementary/cluster_region_plot.csv};
      \addplot+[mark = *, mark size=1pt] table[x=num_clusters, y=num_regions_lns, col sep=comma] {figures/supplementary/cluster_region_plot.csv};
      \addplot+[mark = *, mark size=1pt, color=cyan, mark options={fill=cyan}] table[x=num_clusters, y=num_regions_ddseg1, col sep=comma] {figures/supplementary/cluster_region_plot.csv};
      \legend{EA \cite{yu_edge-aware_2021}, SLIC \cite{achanta_slic_2012}, SNIC \cite{achanta_superpixels_2017},LNS \cite{zhu_learning_2021}, Ours}
    \end{groupplot}
  \end{tikzpicture}
\caption{Number of Superpixels generated plotted against the cluster number input, \ie the number of desired superpixels.}
\label{fig:cluster_region}
\end{figure*}
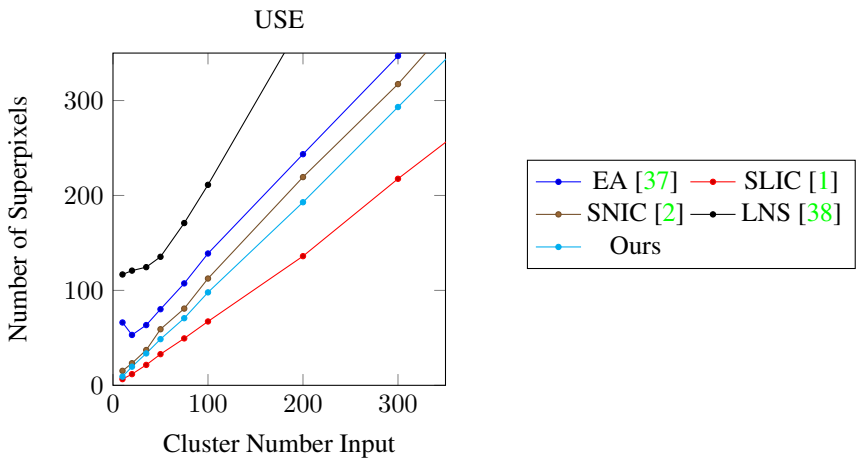

We include further qualitative results to show how our method introduces smooth superpixel boundaries and focuses on prominent image edges while controlling the number of superpixels in a much more stable manner than other unsupervised deep learning based methods such as EA and LNS (\cref{fig:cluster_region}). Before the clustering step we create an evenly spaced pixel grid and choose one cluster center from the $5\times5$ neighbourhood of each pixel on the grid. To keep the grid creation step simple it we choose a $n_w \times n_h$ grid with 
\begin{align}
    n_w &=\left\lfloor \sqrt{\text{Cluster Number Input}\frac{\width}{\height}}))\right\rfloor\\
    n_h &=\left\lfloor \frac{\text{Cluster Number Input}}{n_w}\right\rfloor.
\end{align}
This explains why the number of superpixels produced by our method is 96 for a method input of 100 clusters. To have even stricter control over the number of superpixels we could easily add $\text{Cluster Number Input}-n_w n_h$ other random pixels to our cluster initialization. We decided against it out of simplicity and reproducibility reasons.

\begin{figure*}[h]
    \centering
    
    \begin{subfigure}[c]{0.02\textwidth}
        \rotatebox{90}{10 Clusters}
    \end{subfigure}
    \hfill
    \begin{subfigure}[c]{0.19\linewidth}
        \includegraphics[width=1.0\linewidth]{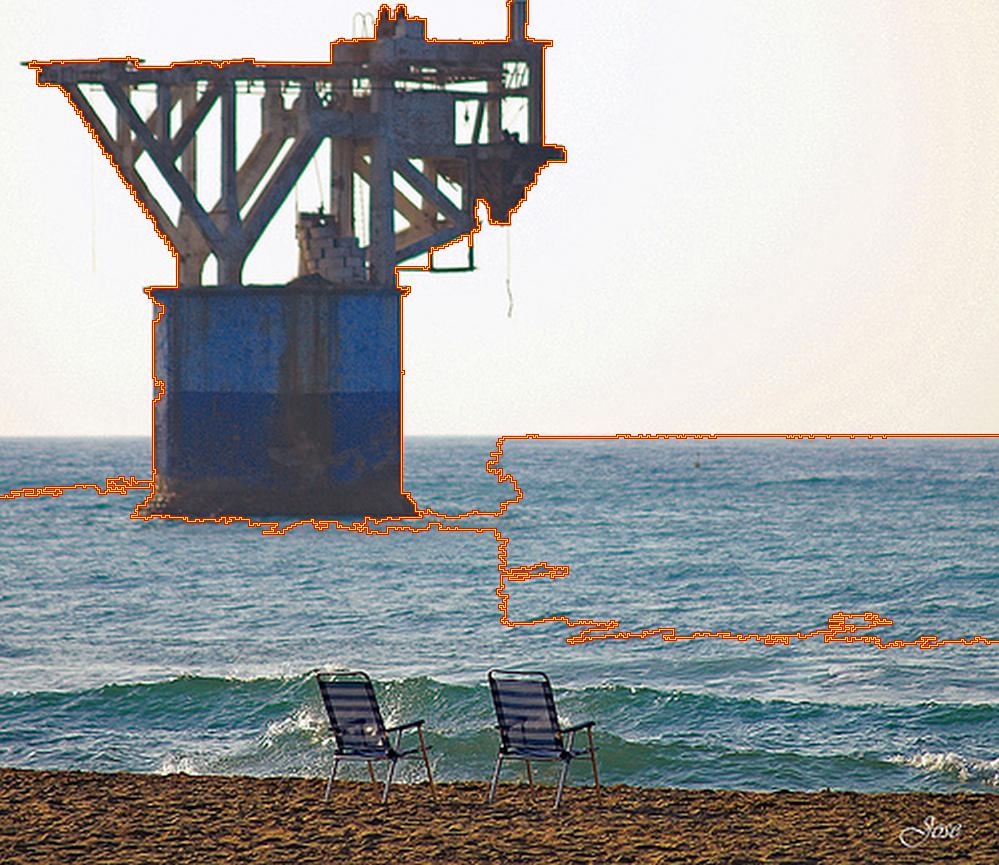}
        \caption{SLIC: 3 SP}
        \label{fig:voc-10-slic}
    \end{subfigure}
    \hfill
    \begin{subfigure}[c]{0.19\linewidth}
        \includegraphics[width=1.0\linewidth]{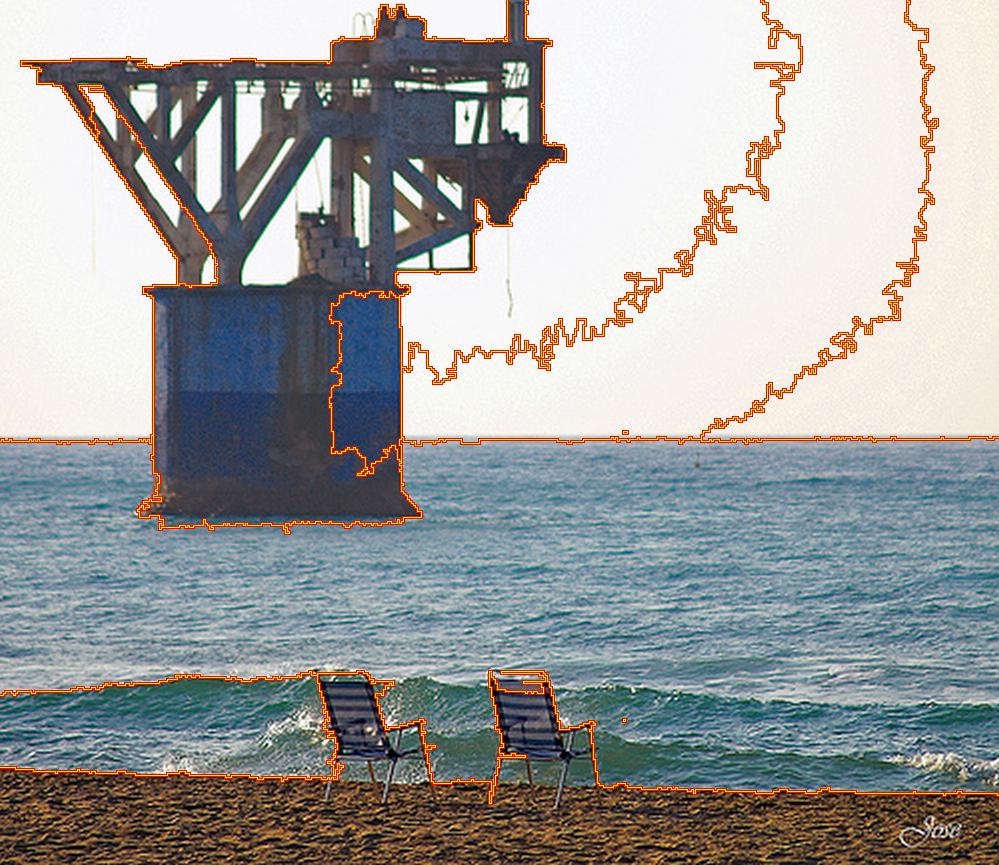}
        \caption{SNIC: 15 SP}
        \label{fig:voc-10-snic}
    \end{subfigure}
    \hfill
    \begin{subfigure}[c]{0.19\linewidth}
        \includegraphics[width=1.0\linewidth]{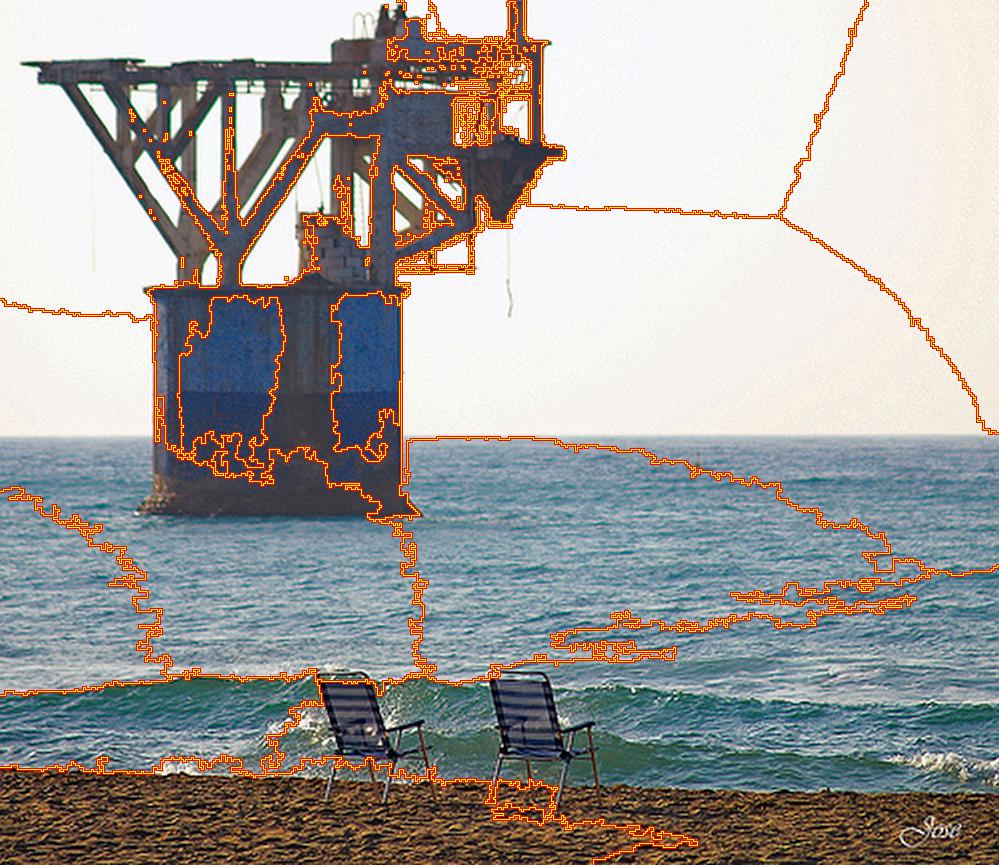}
        \caption{LNS: 48 SP}
        \label{fig:voc-10-lns}
    \end{subfigure}
    \hfill
    \begin{subfigure}[c]{0.19\linewidth}
        \includegraphics[width=1.0\linewidth]{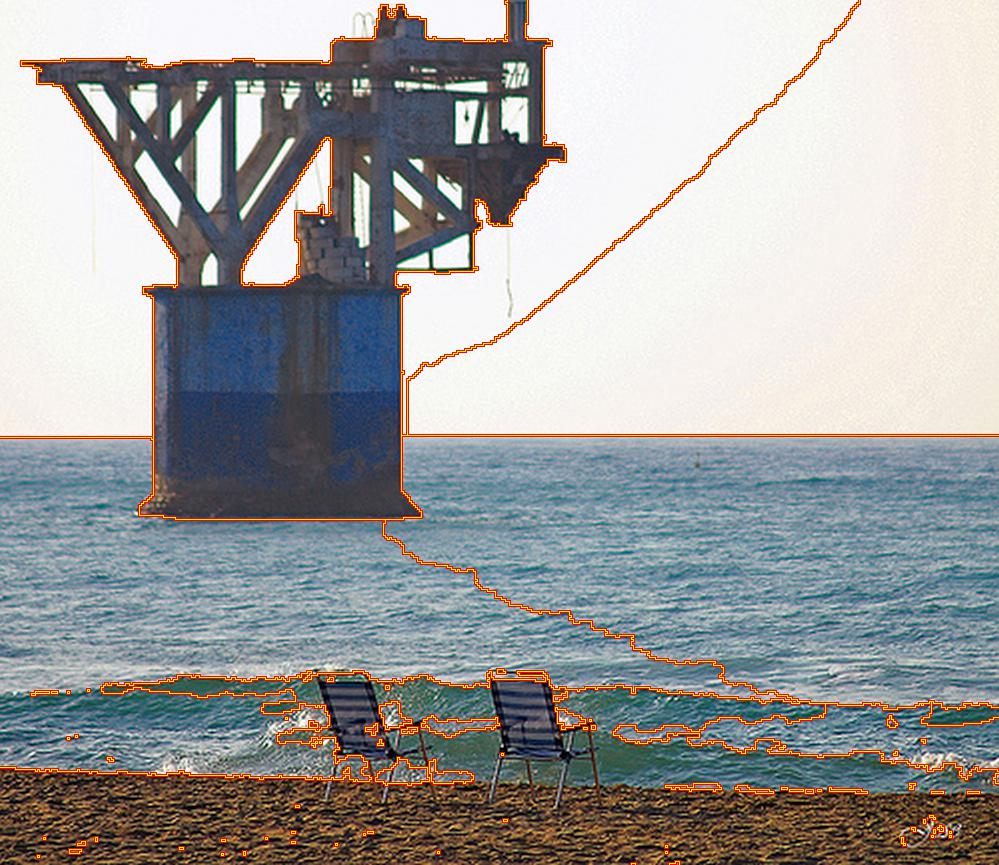}
        \caption{EA: 101 SP}
        \label{fig:voc-10-ea}
    \end{subfigure}
    \hfill
    \begin{subfigure}[c]{0.19\linewidth}
        \includegraphics[width=1.0\linewidth]{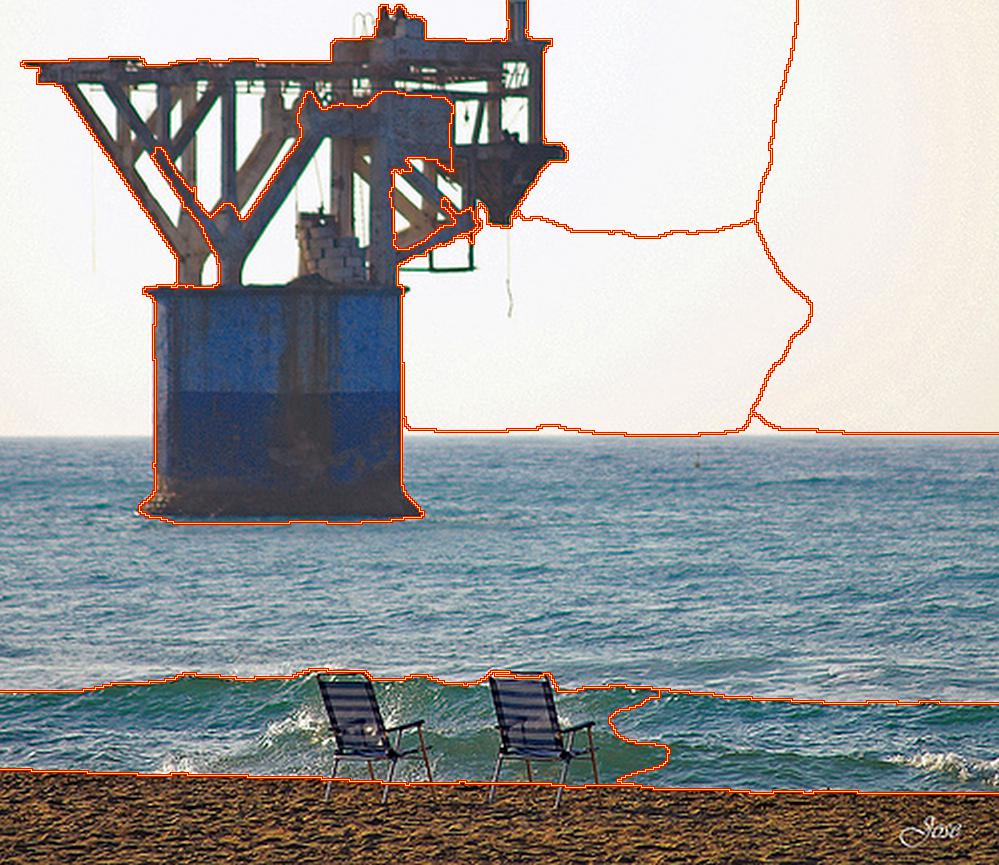}
        \caption{Ours: 10 SP}
        \label{fig:voc-10-ours}
    \end{subfigure}


    \begin{subfigure}[c]{0.02\textwidth}
        \rotatebox{90}{100 Clusters}
    \end{subfigure}
    \hfill
    \begin{subfigure}[c]{0.19\linewidth}
        \includegraphics[width=1.0\linewidth]{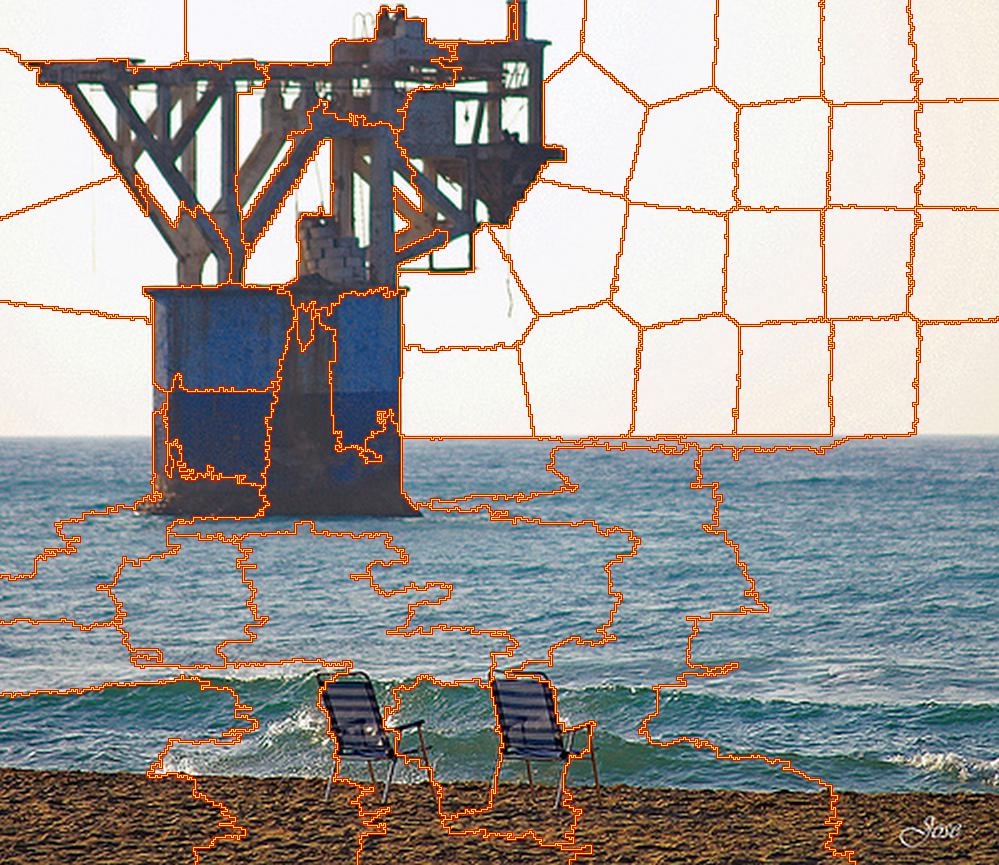}
        \caption{SLIC: 48 SP}
        \label{fig:voc-100-slic}
    \end{subfigure}
    \hfill
    \begin{subfigure}[c]{0.19\linewidth}
        \includegraphics[width=1.0\linewidth]{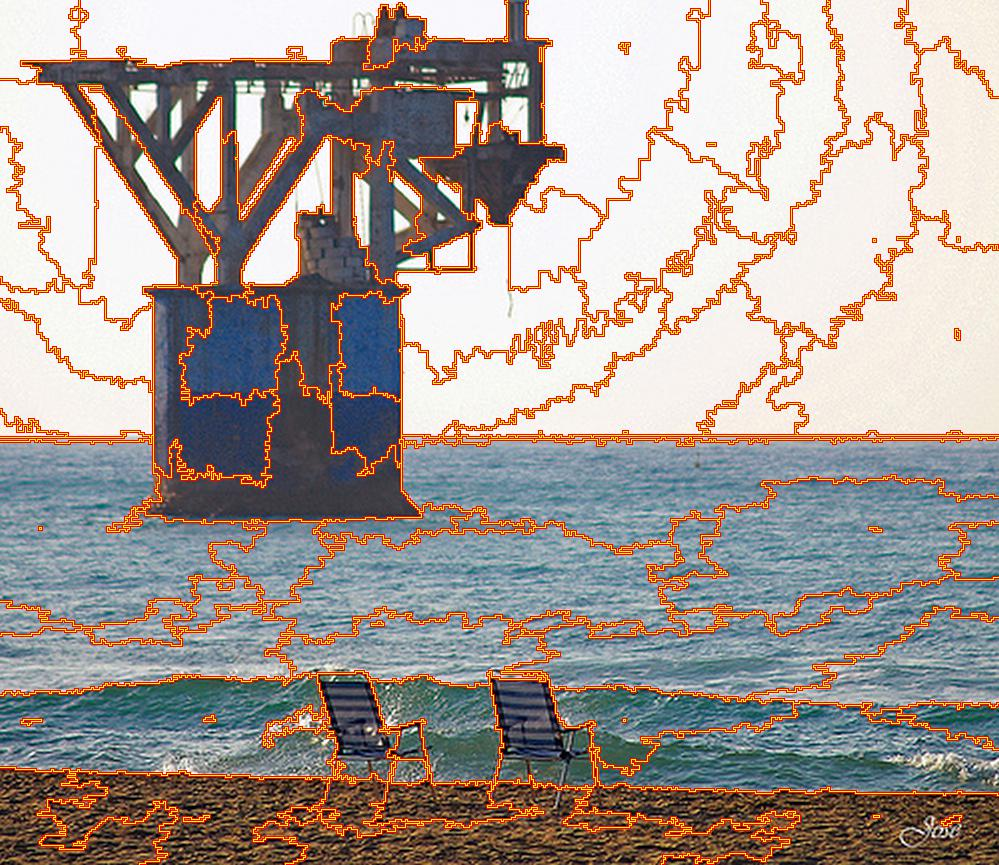}
        \caption{SNIC: 105 SP}
        \label{fig:voc-100-snic}
    \end{subfigure}
    \hfill
    \begin{subfigure}[c]{0.19\linewidth}
        \includegraphics[width=1.0\linewidth]{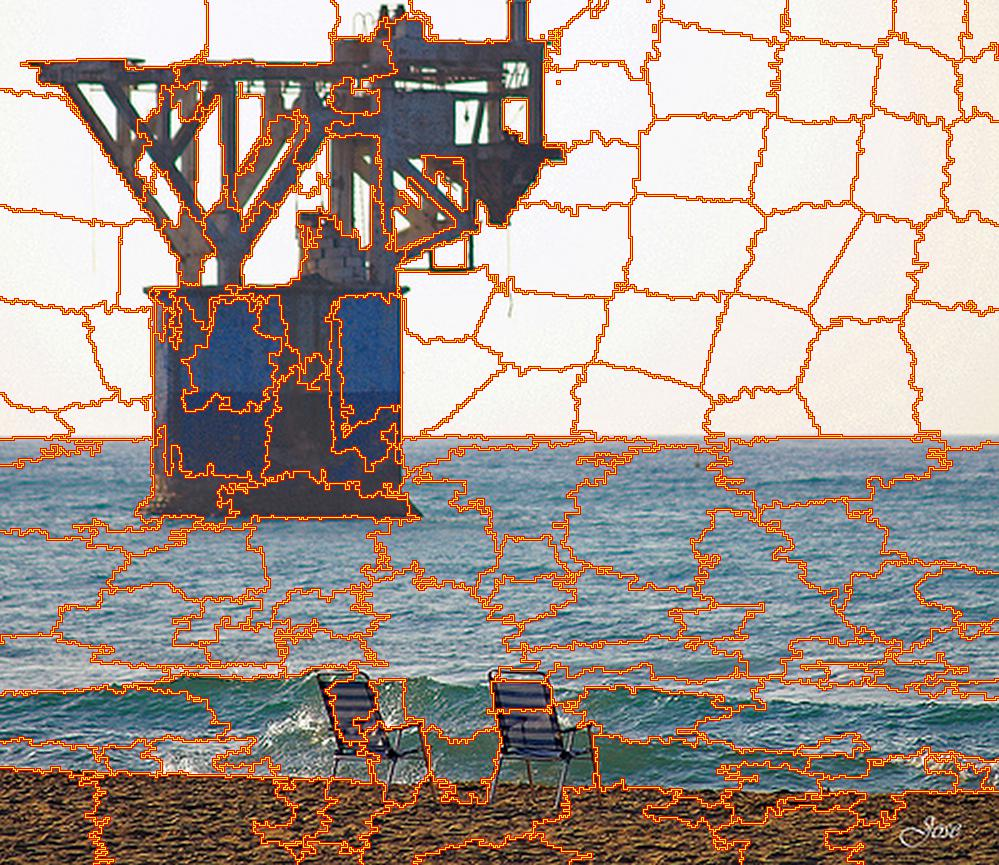}
        \caption{LNS: 222 SP}
        \label{fig:voc-100-lns}
    \end{subfigure}
    \hfill
    \begin{subfigure}[c]{0.19\linewidth}
        \includegraphics[width=1.0\linewidth]{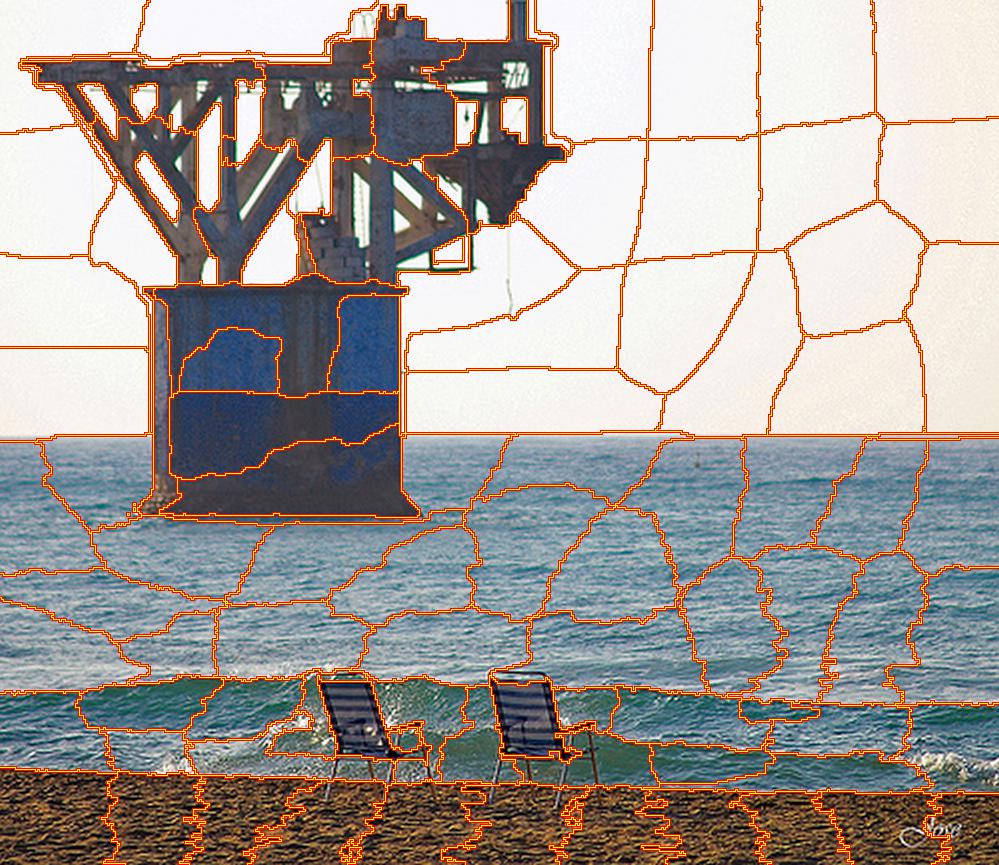}
        \caption{EA: 145 SP}
        \label{fig:voc-100-ea}
    \end{subfigure}
    \hfill
    \begin{subfigure}[c]{0.19\linewidth}
        \includegraphics[width=1.0\linewidth]{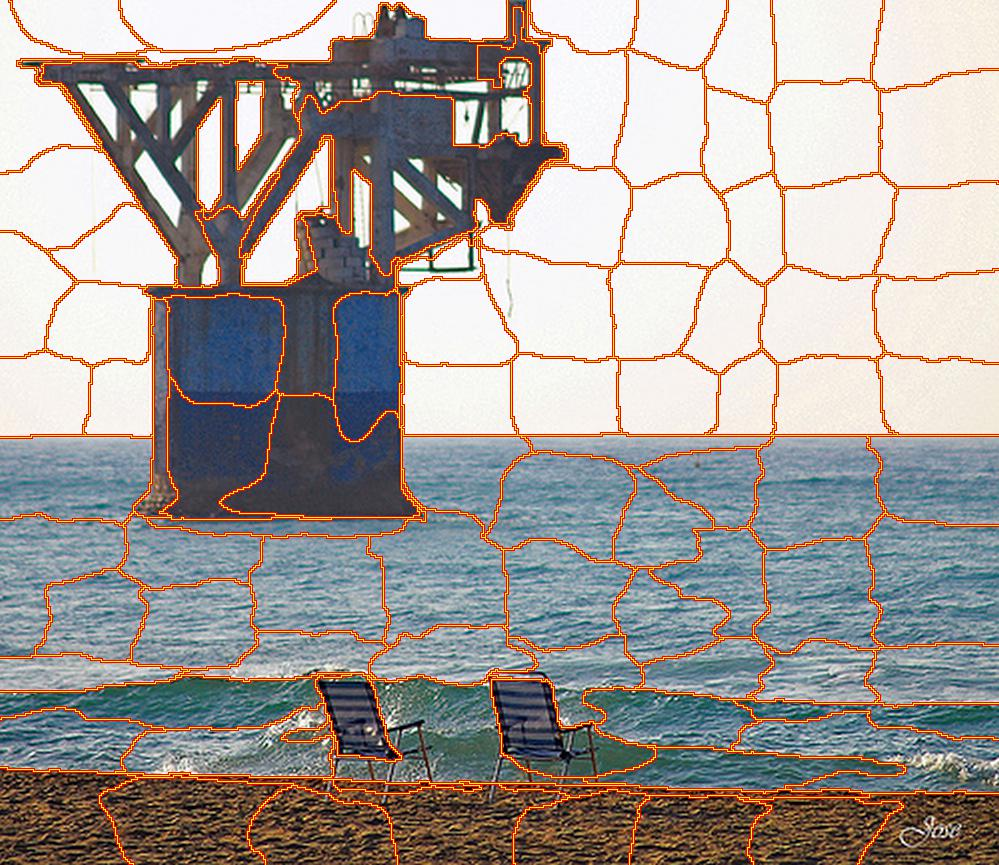}
        \caption{Ours: 96 SP}
        \label{fig:voc-100-ours}
    \end{subfigure}

    \begin{subfigure}[c]{0.02\textwidth}
        \rotatebox{90}{400 Clusters}
    \end{subfigure}
    \hfill
    \begin{subfigure}[c]{0.19\linewidth}
        \includegraphics[width=1.0\linewidth]{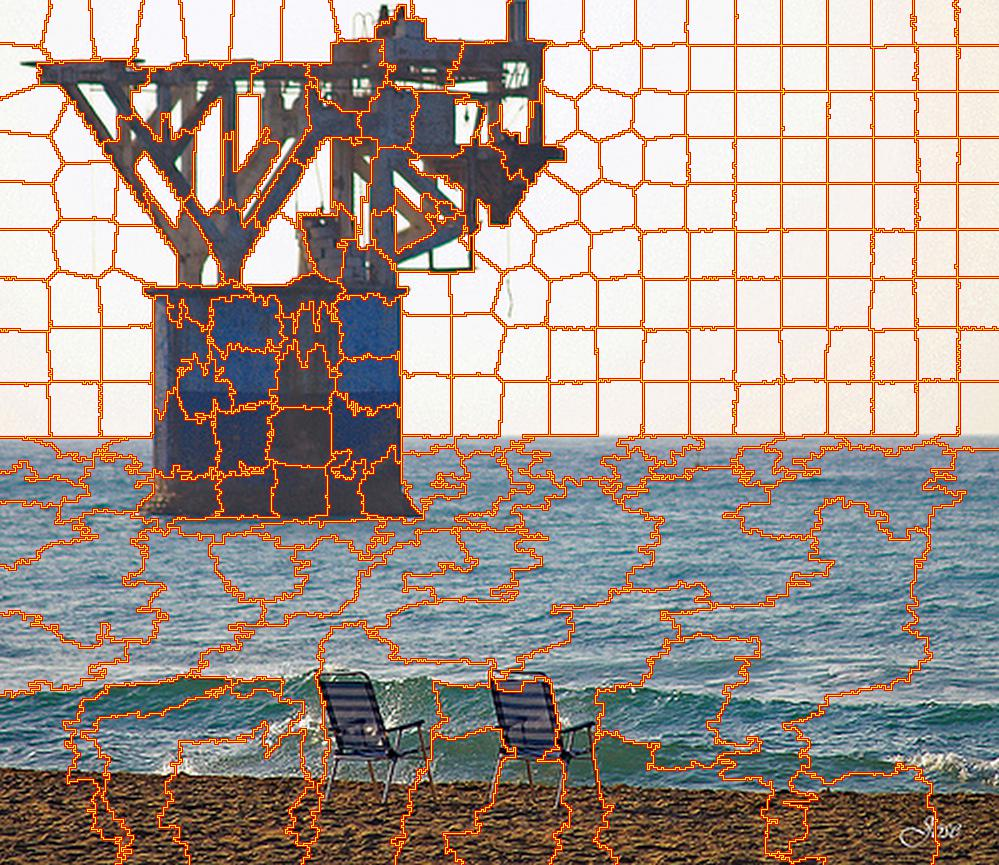}
        \caption{SLIC: 205 SP}
        \label{fig:voc-400-slic}
    \end{subfigure}
    \hfill
    \begin{subfigure}[c]{0.19\linewidth}
        \includegraphics[width=1.0\linewidth]{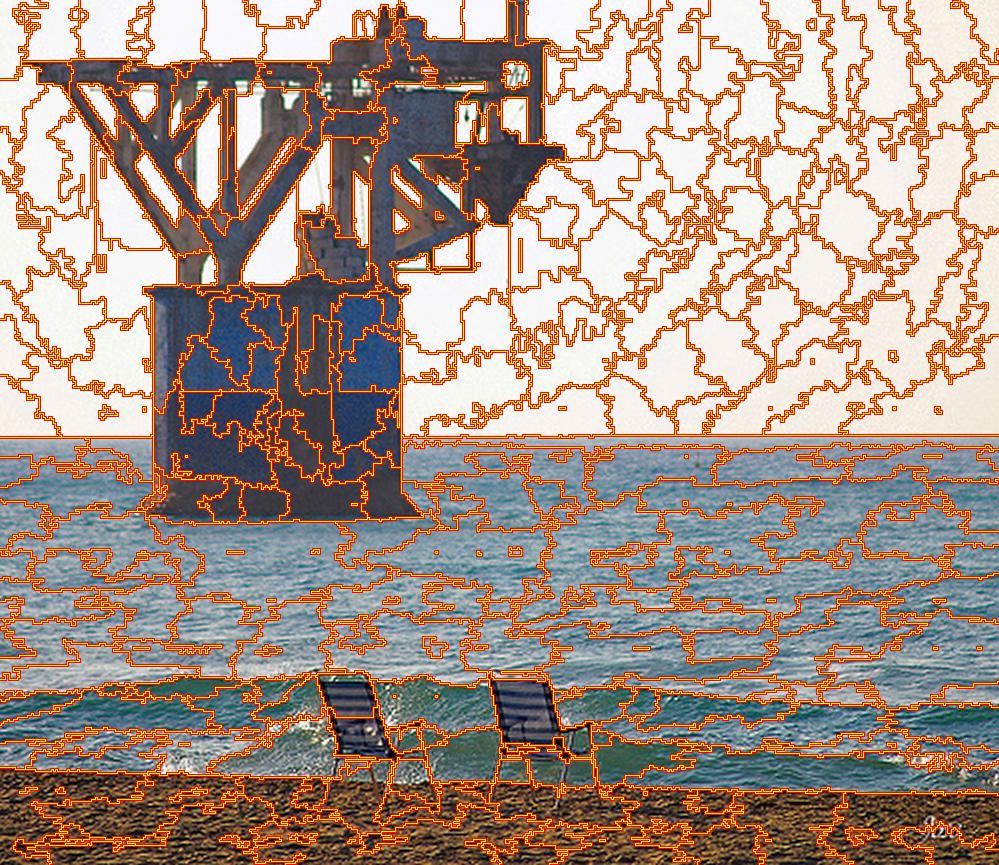}
        \caption{SNIC: 420 SP}
        \label{fig:voc-400-snic}
    \end{subfigure}
    \hfill
    \begin{subfigure}[c]{0.19\linewidth}
        \includegraphics[width=1.0\linewidth]{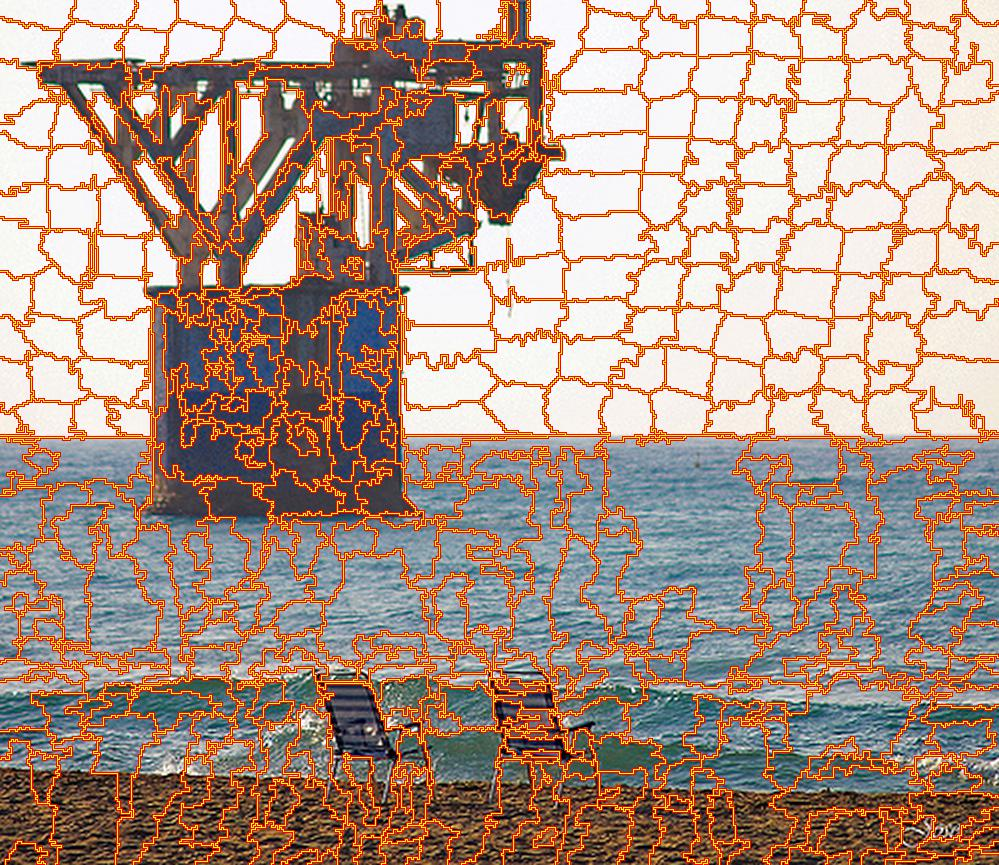}
        \caption{LNS: 782 SP}
        \label{fig:voc-400-lns}
    \end{subfigure}
    \hfill
    \begin{subfigure}[c]{0.19\linewidth}
        \includegraphics[width=1.0\linewidth]{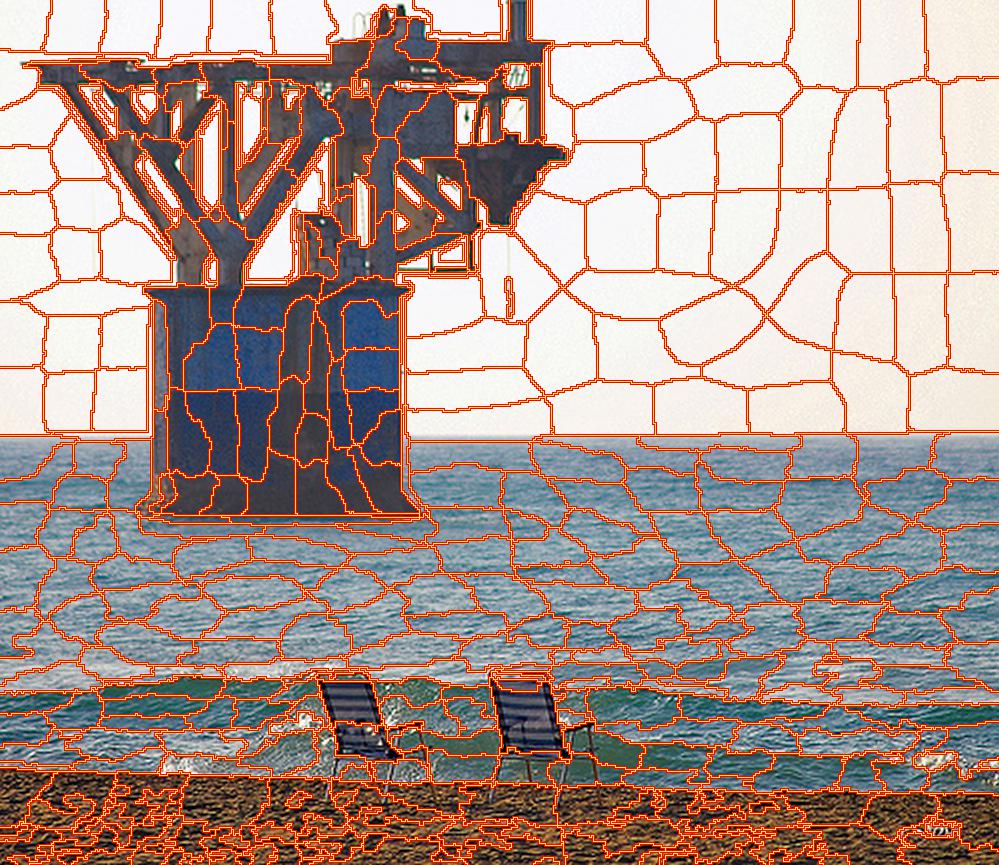}
        \caption{EA: 471 SP}
        \label{fig:voc-400-ea}
    \end{subfigure}
    \hfill
    \begin{subfigure}[c]{0.19\linewidth}
        \includegraphics[width=1.0\linewidth]{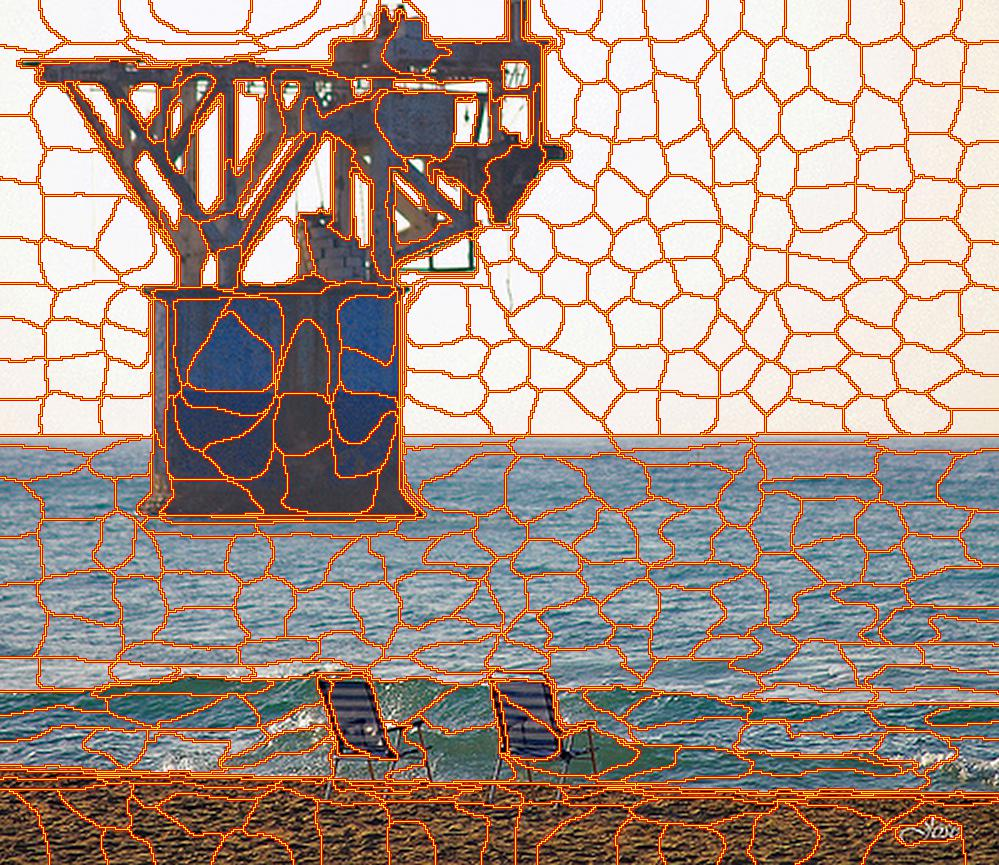}
        \caption{Ours: 400 SP}
        \label{fig:voc-400-ours}
    \end{subfigure}

    \begin{subfigure}[c]{0.19\linewidth}
        \includegraphics[width=1.0\linewidth]{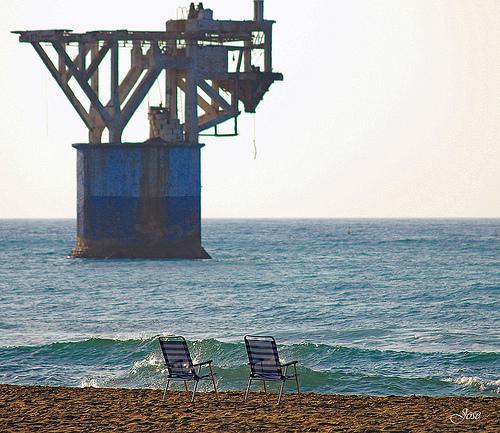}
        \caption{Original Image}
        \label{fig:voc-original}
    \end{subfigure}
    \begin{subfigure}[c]{0.19\linewidth}
        \includegraphics[width=1.0\linewidth]{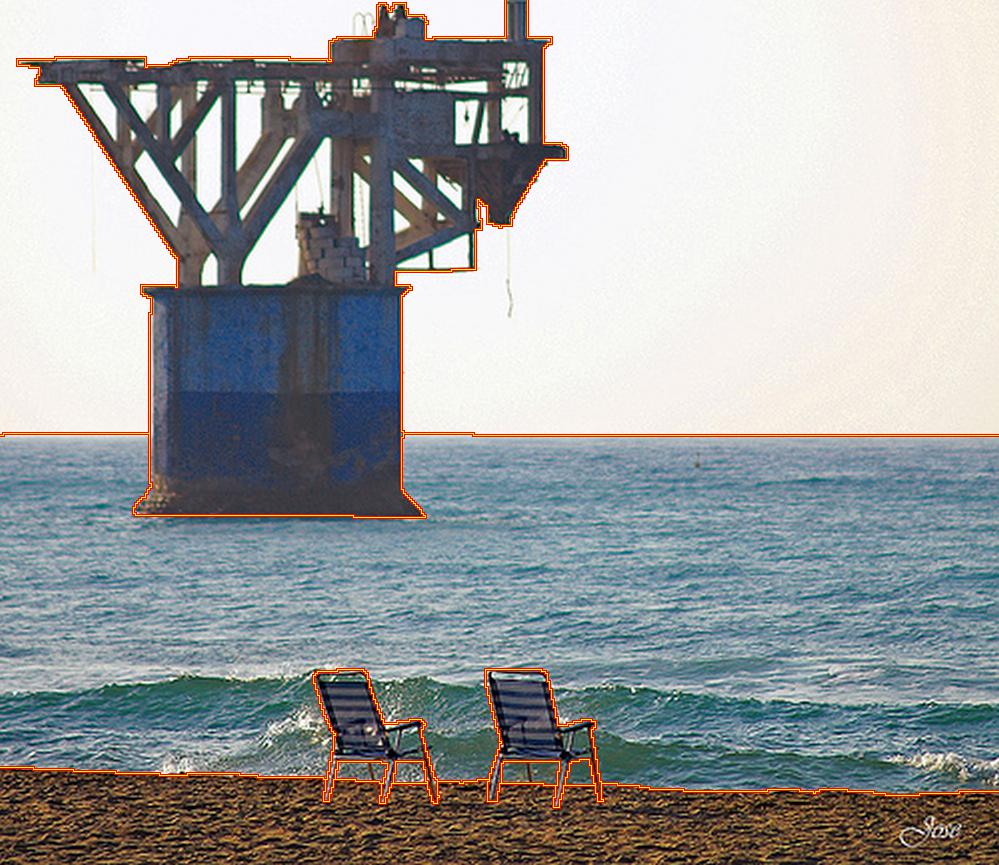}
        \caption{Ground Truth}
        \label{fig:voc-gt}
    \end{subfigure}

    \caption{Qualitative Comparison of Superpixel Algorithms on a sample image from PASCAL-Context. The algorithms were given the cluster parameters 10, 100 and 400 (row-wise) and the produced oversegmentations are shown with the actual number of generated superpixels (SP).}
    \label{fig:qual_voc}
\end{figure*}

\begin{figure*}[h]
    \centering
    
    \begin{subfigure}[c]{0.02\textwidth}
        \rotatebox{90}{10 Clusters}
    \end{subfigure}
    \hfill
    \begin{subfigure}[c]{0.19\linewidth}
        \includegraphics[width=1.0\linewidth]{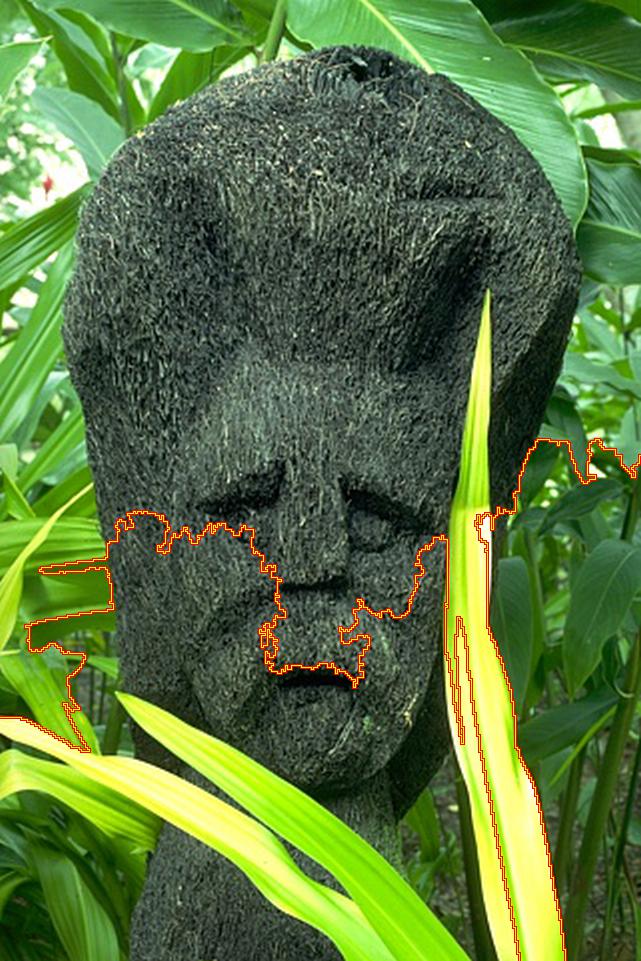}
        \caption{SLIC: 3 SP}
        \label{fig:bsds-10-slic}
    \end{subfigure}
    \hfill
    \begin{subfigure}[c]{0.19\linewidth}
        \includegraphics[width=1.0\linewidth]{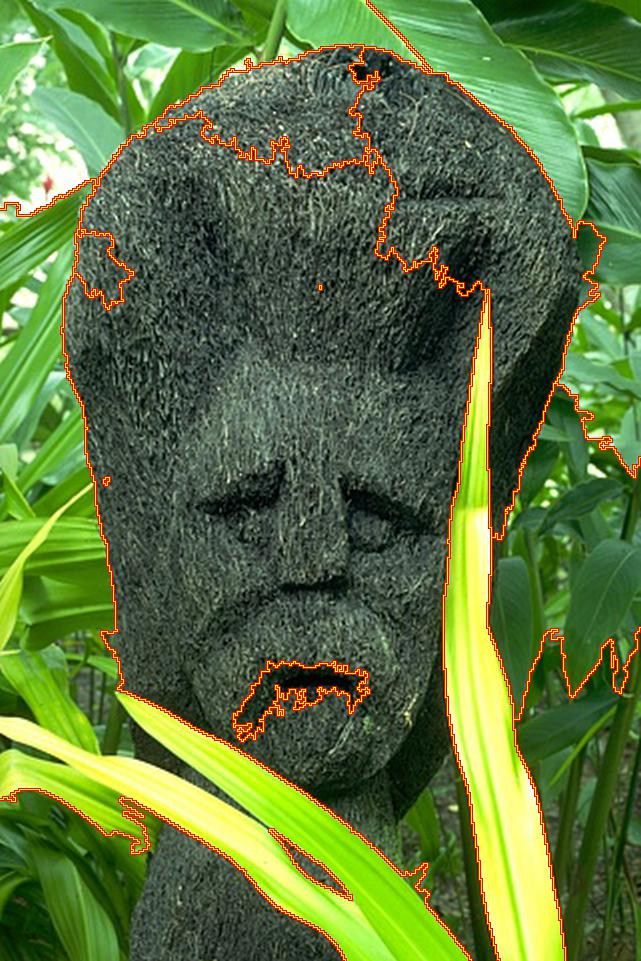}
        \caption{SNIC: 15 SP}
        \label{fig:bsds-10-snic}
    \end{subfigure}
    \hfill
    \begin{subfigure}[c]{0.19\linewidth}
        \includegraphics[width=1.0\linewidth]{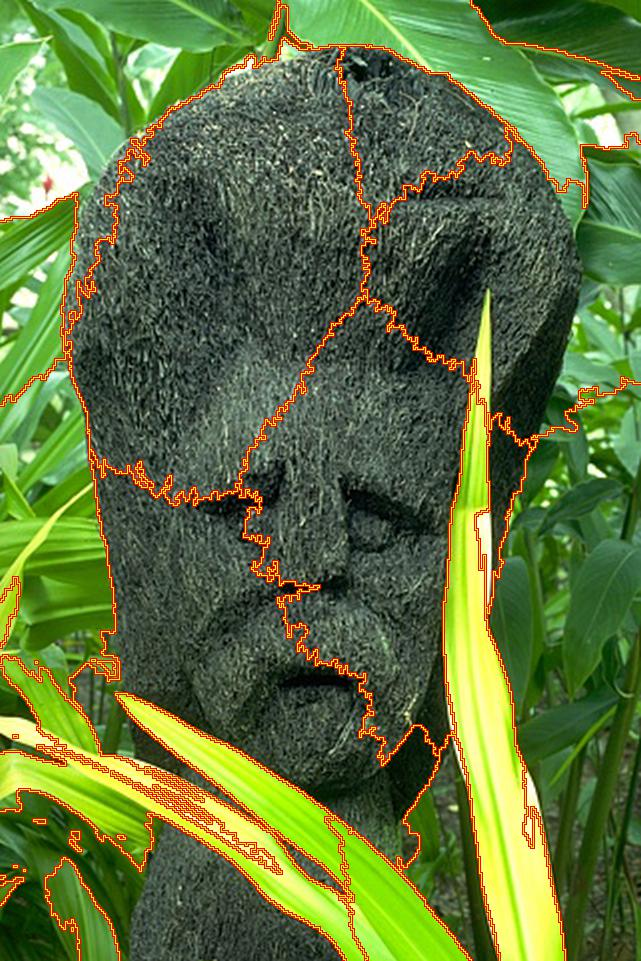}
        \caption{LNS: 48 SP}
        \label{fig:bsds-10-lns}
    \end{subfigure}
    \hfill
    \begin{subfigure}[c]{0.19\linewidth}
        \includegraphics[width=1.0\linewidth]{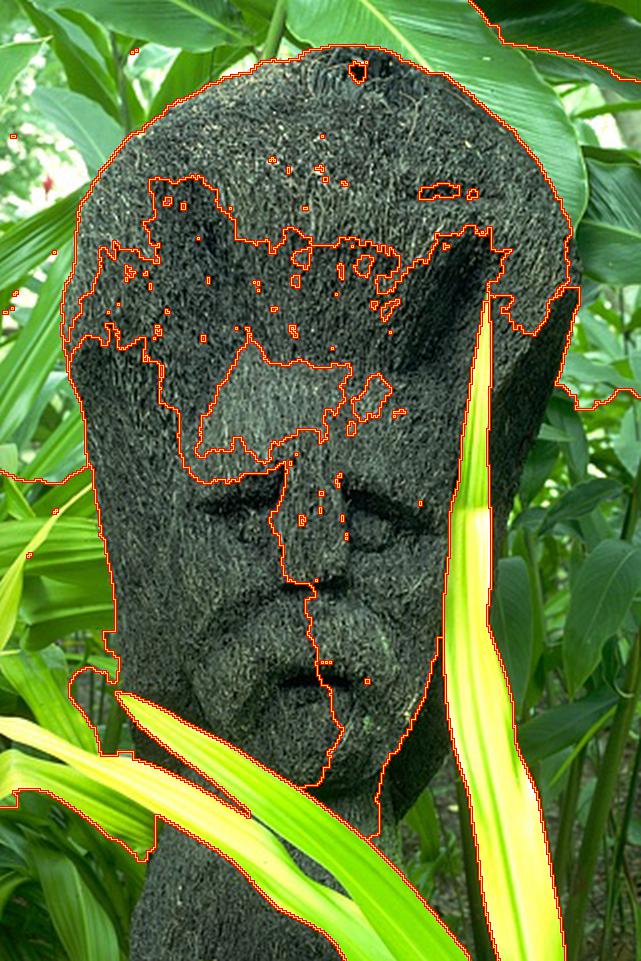}
        \caption{EA: 101 SP}
        \label{fig:bsds-10-ea}
    \end{subfigure}
    \hfill
    \begin{subfigure}[c]{0.19\linewidth}
        \includegraphics[width=1.0\linewidth]{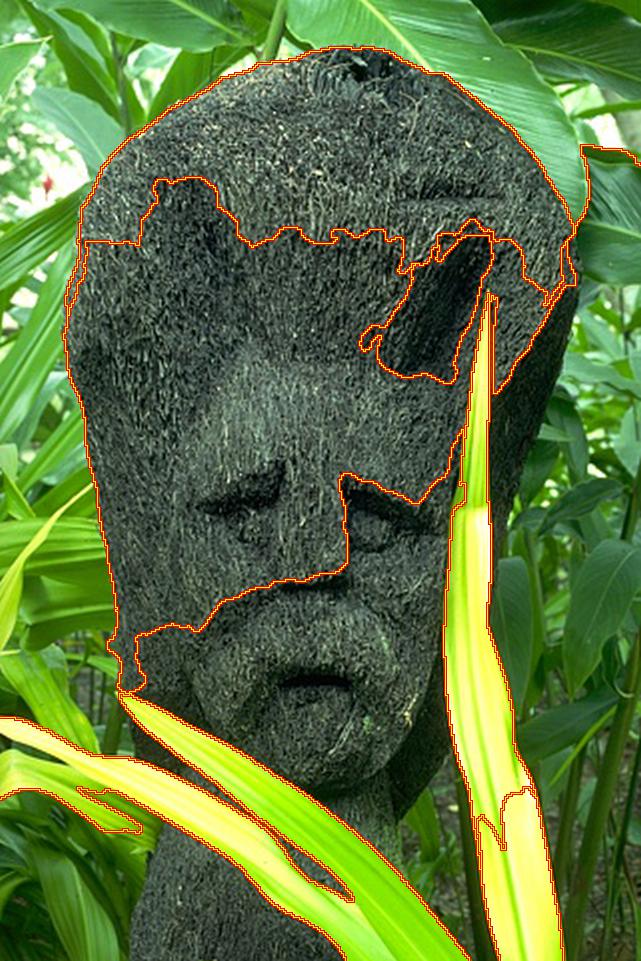}
        \caption{Ours: 10 SP}
        \label{fig:bsds-10-ours}
    \end{subfigure}


    \begin{subfigure}[c]{0.02\textwidth}
        \rotatebox{90}{100 Clusters}
    \end{subfigure}
    \hfill
    \begin{subfigure}[c]{0.19\linewidth}
        \includegraphics[width=1.0\linewidth]{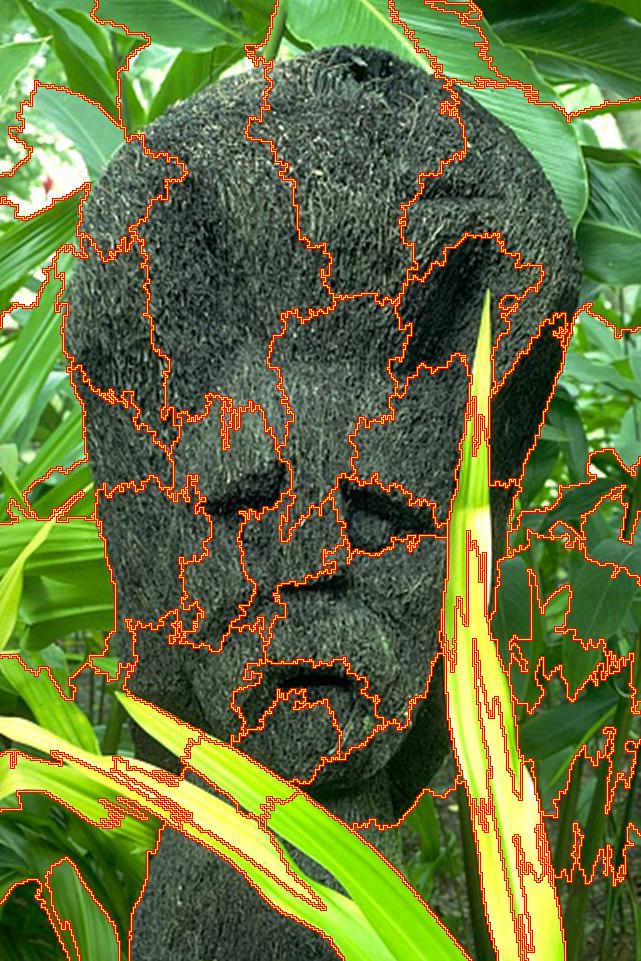}
        \caption{SLIC: 48 SP}
        \label{fig:bsds-100-slic}
    \end{subfigure}
    \hfill
    \begin{subfigure}[c]{0.19\linewidth}
        \includegraphics[width=1.0\linewidth]{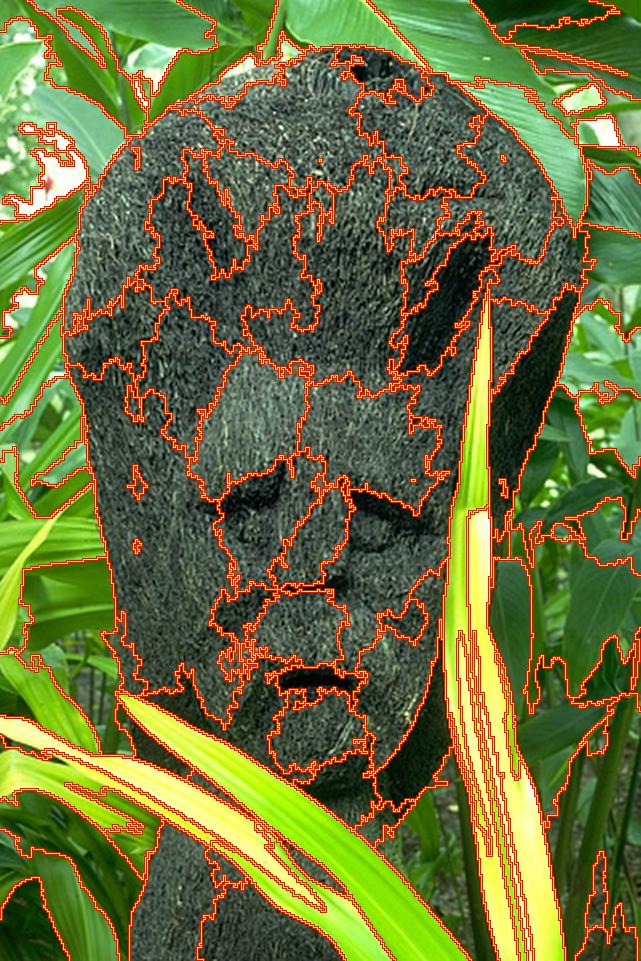}
        \caption{SNIC: 105 SP}
        \label{fig:bsds-100-snic}
    \end{subfigure}
    \hfill
    \begin{subfigure}[c]{0.19\linewidth}
        \includegraphics[width=1.0\linewidth]{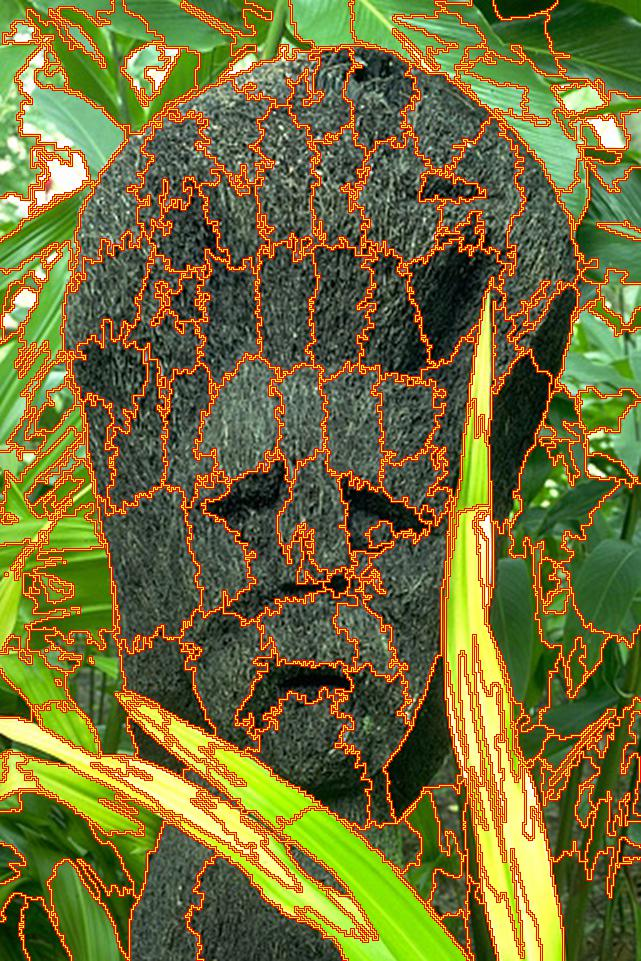}
        \caption{LNS: 222 SP}
        \label{fig:bsds-100-lns}
    \end{subfigure}
    \hfill
    \begin{subfigure}[c]{0.19\linewidth}
        \includegraphics[width=1.0\linewidth]{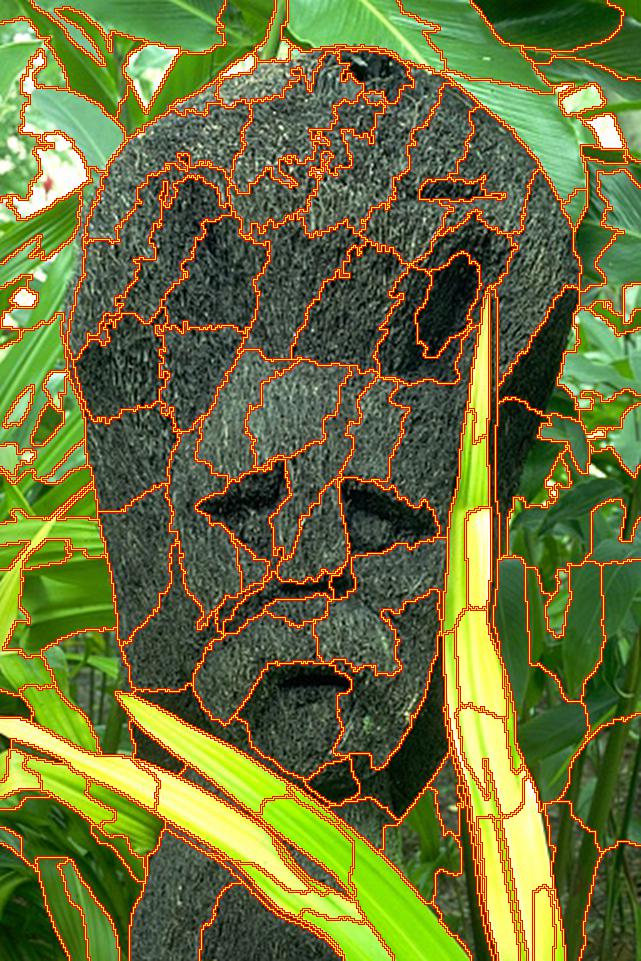}
        \caption{EA: 145 SP}
        \label{fig:bsds-100-ea}
    \end{subfigure}
    \hfill
    \begin{subfigure}[c]{0.19\linewidth}
        \includegraphics[width=1.0\linewidth]{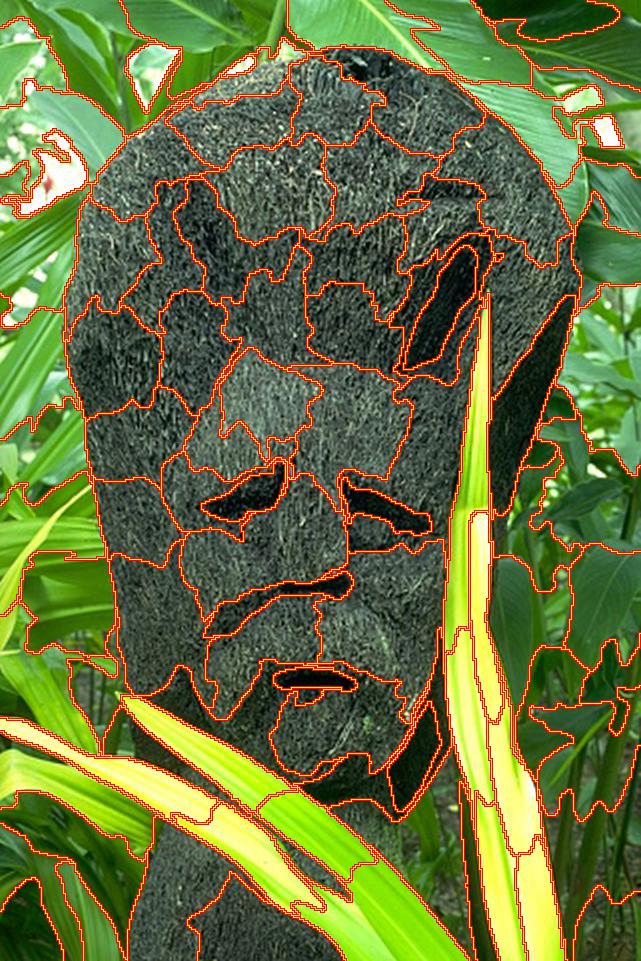}
        \caption{Ours: 96 SP}
        \label{fig:bsds-100-ours}
    \end{subfigure}

    \begin{subfigure}[c]{0.02\textwidth}
        \rotatebox{90}{400 Clusters}
    \end{subfigure}
    \hfill
    \begin{subfigure}[c]{0.19\linewidth}
        \includegraphics[width=1.0\linewidth]{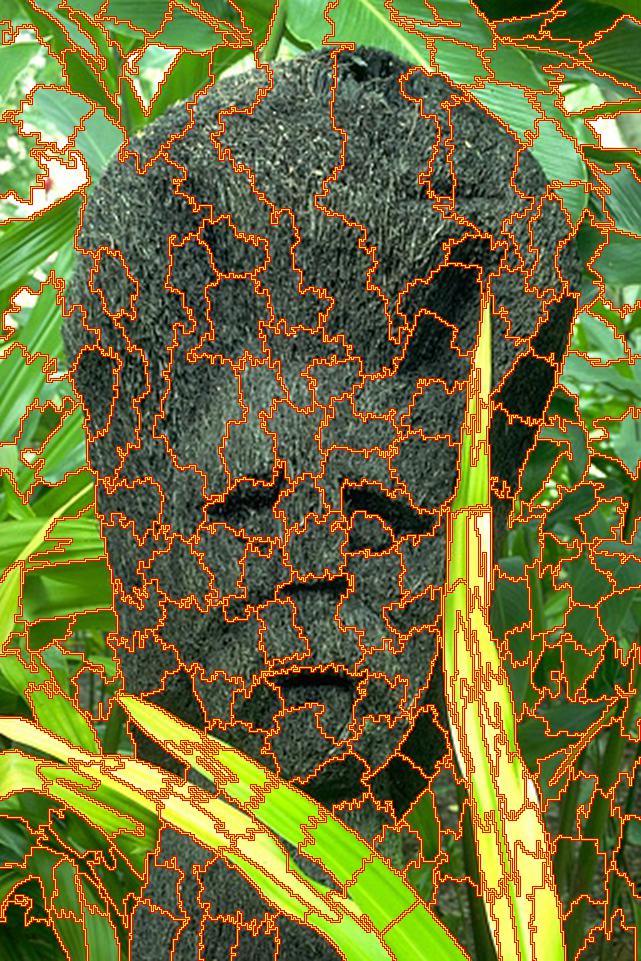}
        \caption{SLIC: 205 SP}
        \label{fig:bsds-400-slic}
    \end{subfigure}
    \hfill
    \begin{subfigure}[c]{0.19\linewidth}
        \includegraphics[width=1.0\linewidth]{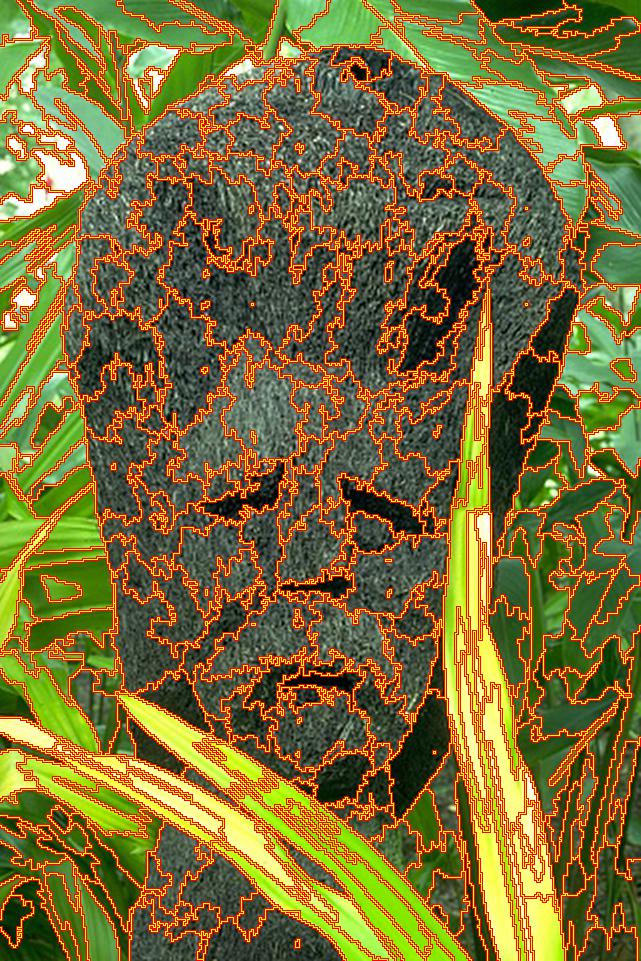}
        \caption{SNIC: 420 SP}
        \label{fig:bsds-400-snic}
    \end{subfigure}
    \hfill
    \begin{subfigure}[c]{0.19\linewidth}
        \includegraphics[width=1.0\linewidth]{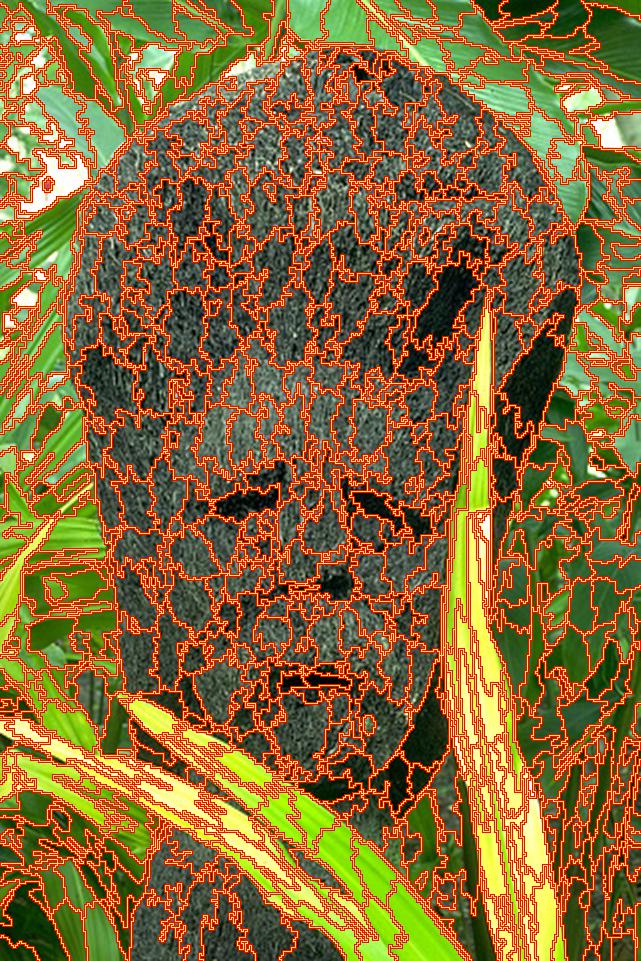}
        \caption{LNS: 782 SP}
        \label{fig:bsds-400-lns}
    \end{subfigure}
    \hfill
    \begin{subfigure}[c]{0.19\linewidth}
        \includegraphics[width=1.0\linewidth]{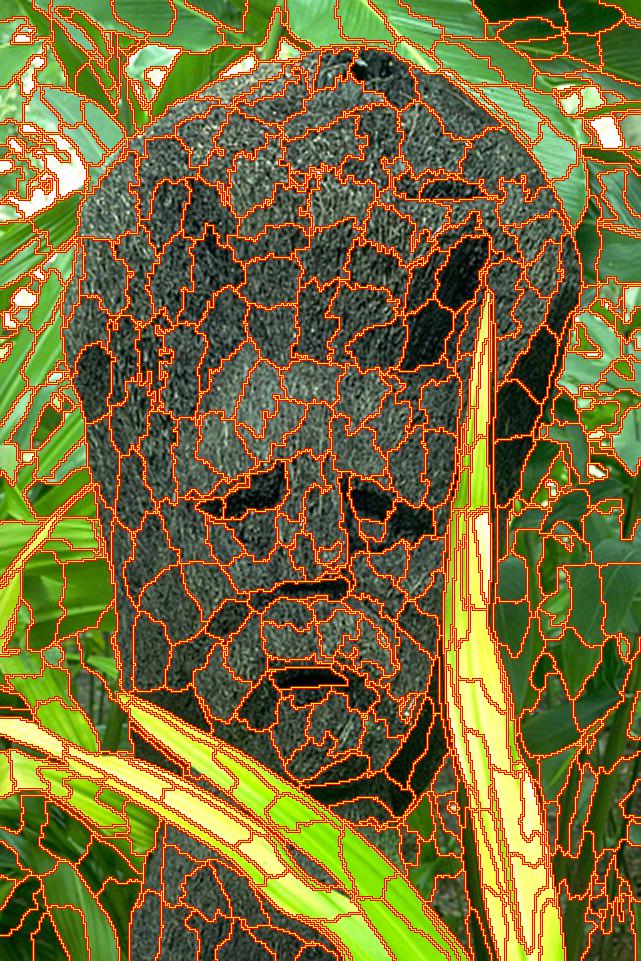}
        \caption{EA: 471 SP}
        \label{fig:bsds-400-ea}
    \end{subfigure}
    \hfill
    \begin{subfigure}[c]{0.19\linewidth}
        \includegraphics[width=1.0\linewidth]{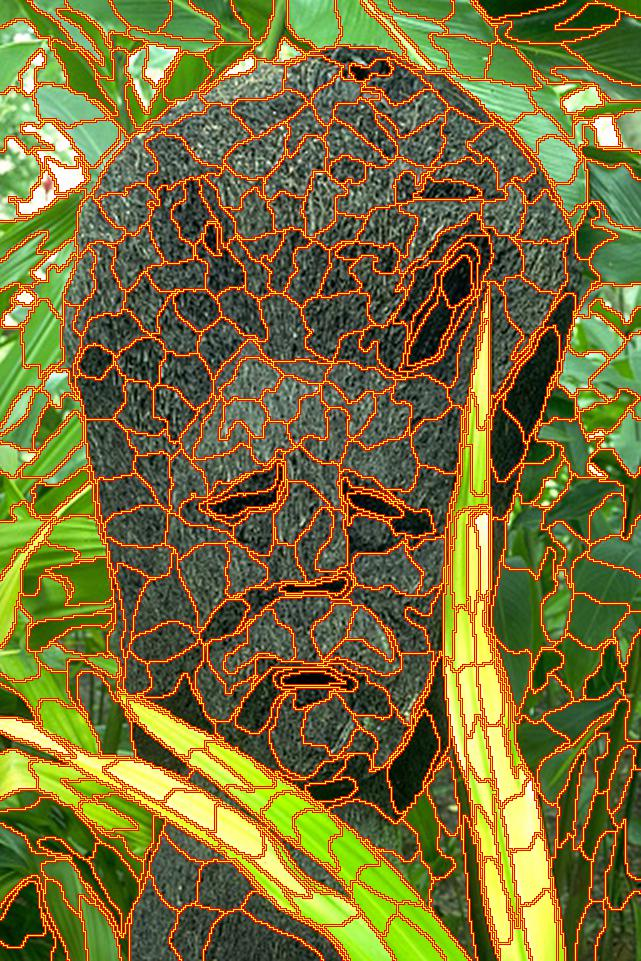}
        \caption{Ours: 400 SP}
        \label{fig:bsds-400-ours}
    \end{subfigure}

    \begin{subfigure}[c]{0.02\textwidth}
        \rotatebox{90}{Ground Truths}
    \end{subfigure}
    \hfill
    \begin{subfigure}[c]{0.19\linewidth}
        \includegraphics[width=1.0\linewidth]{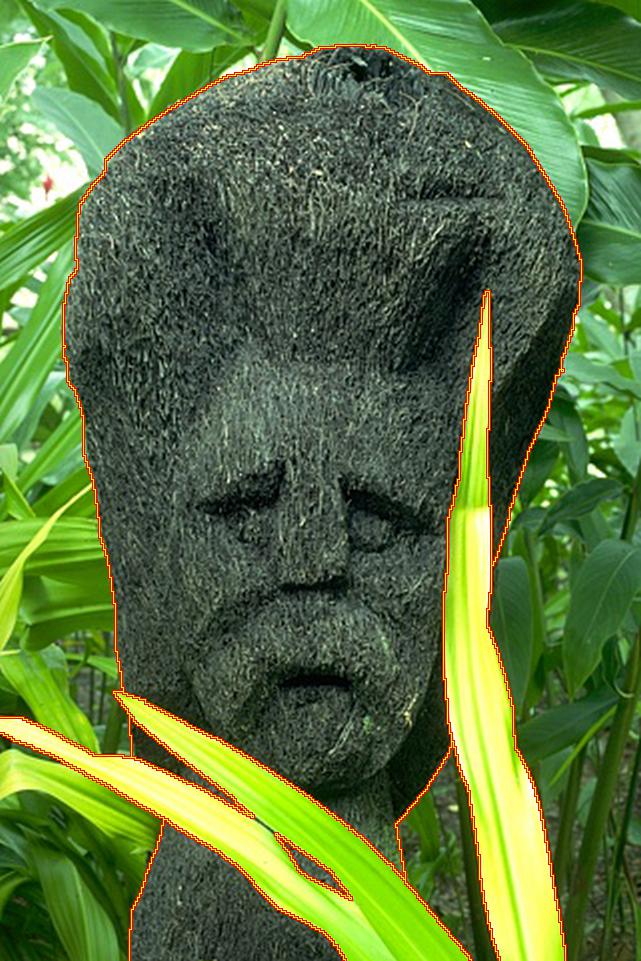}
        \caption{Ground Truth 1}
        \label{fig:bsds-gt1}
    \end{subfigure}
    \hfill
    \begin{subfigure}[c]{0.19\linewidth}
        \includegraphics[width=1.0\linewidth]{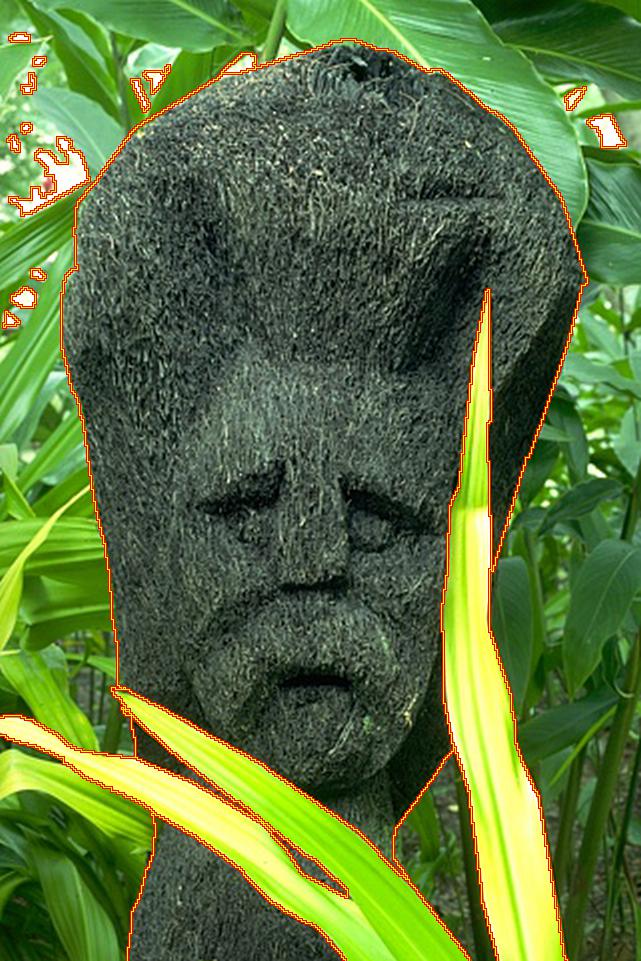}
        \caption{Ground Truth 2}
        \label{fig:bsds-gt2}
    \end{subfigure}
    \hfill
    \begin{subfigure}[c]{0.19\linewidth}
        \includegraphics[width=1.0\linewidth]{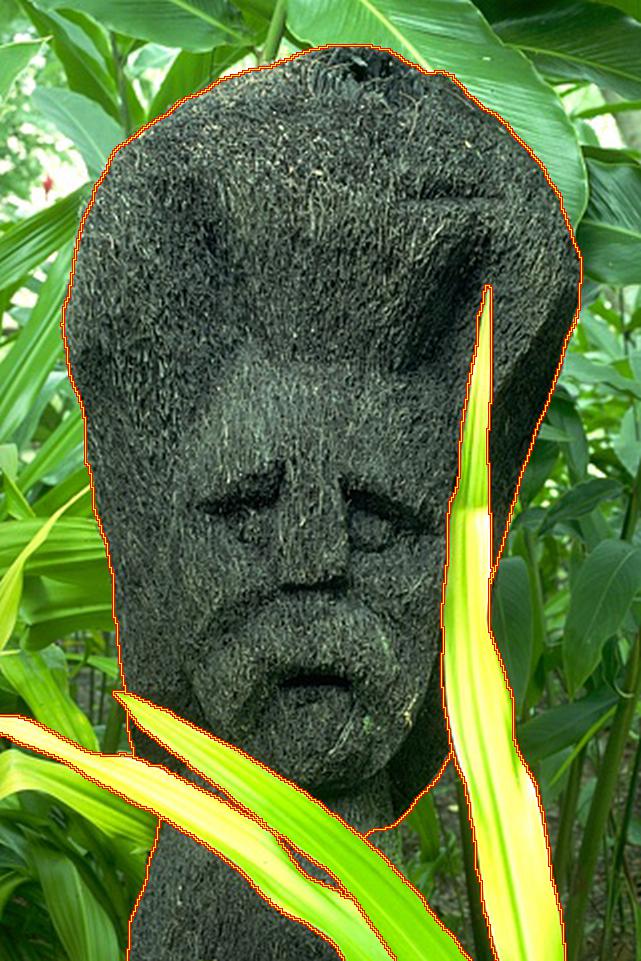}
        \caption{Ground Truth 3}
        \label{fig:bsds-gt3}
    \end{subfigure}
    \hfill
    \begin{subfigure}[c]{0.19\linewidth}
        \includegraphics[width=1.0\linewidth]{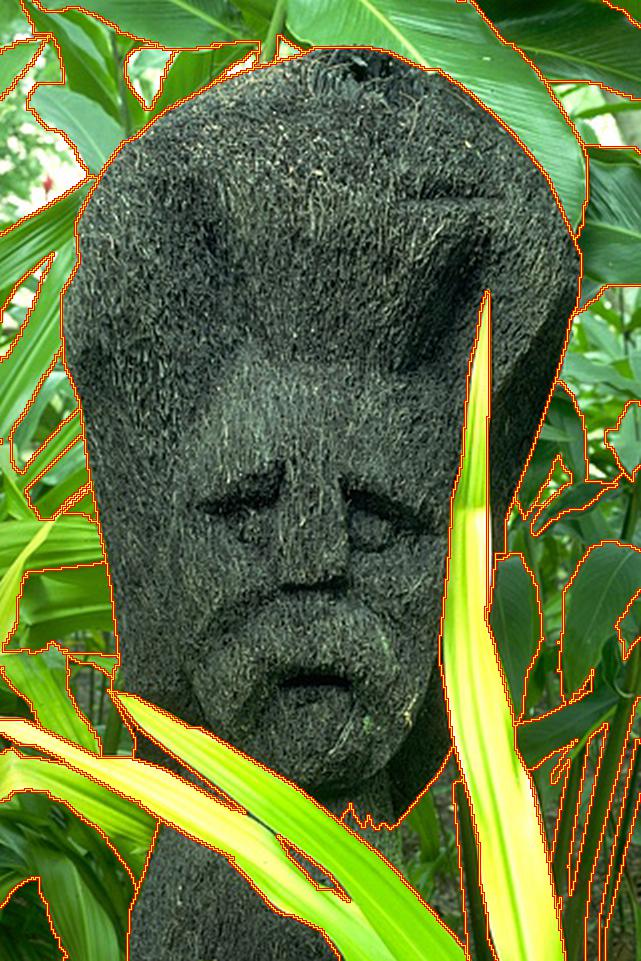}
        \caption{Ground Truth 4}
        \label{fig:bsds-gt4}
    \end{subfigure}
    \hfill
    \begin{subfigure}[c]{0.19\linewidth}
        \includegraphics[width=1.0\linewidth]{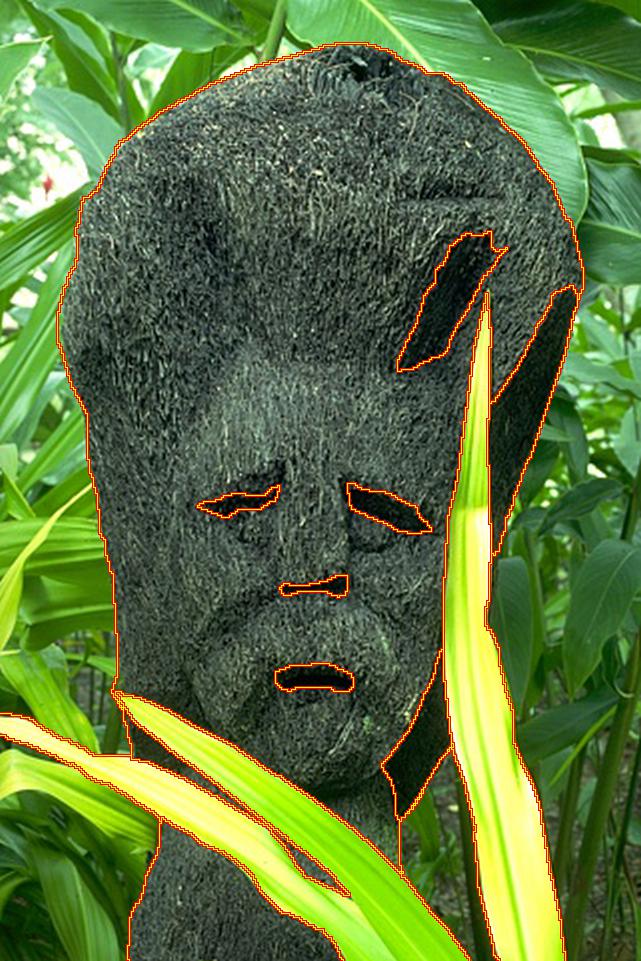}
        \caption{Ground Truth 5}
        \label{fig:bsds-gt5}
    \end{subfigure}

    \caption{Qualitative Comparison of Superpixel Algorithms on a sample image from BSDS 500. The algorithms were given the cluster parameters 10, 100 and 400 (row-wise) and the produced oversegmentations are shown with the actual number of generated superpixels (SP).}
    \label{fig:qual_bsds}
\end{figure*}

In \Cref{fig:qual_voc} we can see superpixel partitions of a sample image of the PASCAL-Context dataset. LNS and EA generate more superpixels than desired when their partitions are focused on seemingly arbitrary edges in the water and tower, respectively. This results in spatially disconnected clusters and 1-pixel wide superpixels around contrasts. In both figures 1-pixel wide superpixels cannot be identified as such and only make the edge marker appear thicker. In \Cref{fig:qual_bsds} we can see how our method captures the boundaries of the bright leaves, as well as the top part of the statue with just 10 superpixels. In general the superpixel boundaries produced by our method are smoother and less distracted by small edges in the statue's texture. We can also observe an example of how the blurred deep decoder's line continuity enforcement results in better edge adherence between the two prominent leaves at the bottom of the image (\cref{fig:bsds-10-ours}) in comparison to EA (\cref{fig:bsds-10-ea}).


\clearpage
\section{Network Under-Parameterization and Input Blurring}

\begin{figure*}[h]
    \centering
    \begin{subfigure}[c]{0.19\linewidth}
        \includegraphics[width=1.0\linewidth]{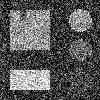}
        \caption{Noisy Image}
        \label{fig:noisy}
    \end{subfigure}
    \hfill
    \begin{subfigure}[c]{0.19\linewidth}
        \includegraphics[width=1.0\linewidth]{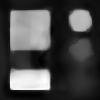}
        \caption{Tiny Decoder - No Blur}
        \label{fig:tiny}
    \end{subfigure}
    \hfill
    \begin{subfigure}[c]{0.19\linewidth}
        \includegraphics[width=1.0\linewidth]{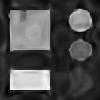}
        \caption{Decoder - No Blur}
        \label{fig:noblur}
    \end{subfigure}
    \hfill
    \begin{subfigure}[c]{0.19\linewidth}
        \includegraphics[width=1.0\linewidth]{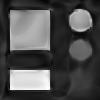}
        \caption{Decoder - Blur}
        \label{fig:blur}
    \end{subfigure}
    \hfill
    \begin{subfigure}[c]{0.19\linewidth}
        \includegraphics[width=1.0\linewidth]{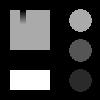}
        \caption{Noise-free}
        \label{fig:original}
    \end{subfigure}
      
    \caption{We reconstruct the noisy image (\cref{fig:noisy}) with a highly under-parameterized deep decoder (\cref{fig:tiny}) and a larger decoder, without (\cref{fig:noblur}) and with blurred (\cref{fig:blur}) network input. The highly under-parameterized decoder has 4 layers with 8 channels and a $3\times3$-sized network input. The larger decoder consists of 4 layers with 32 channels and a $3\times3$-sized network input. The network input was blurred with a Gaussian kernel of standard deviation $\blurstd=5$.}
    \label{fig:artificial}
\end{figure*}

In this section the effect of blurring the network input is further investigated. As we reduce the number of layers, the number of hidden channels and the input size we can increase the denoising effect of the deep decoder. We want to discuss how network under-parameterization and input blurring have complimentary effects. Under the assumption that the noisy image's underlying objects have smooth textures a Deep Decoder needs to be highly under-parameterized in order to produce smooth regions. This however, reduces the deep decoder's ability to correctly reproduce the shapes of the objects in the original image (\cref{fig:tiny}). In \Cref{fig:noblur} it is visualized how a larger decoder without blurred input is able to reconstruct the object's shapes but introduces small artifacts within the shapes and in the background. By blurring the input of this larger decoder we can enforce smooth regions without limiting the Deep Decoder's ability to reproduce shapes and reduce reconstruction artifacts. Furthermore, edge connectivity is enforced, as we can see in \Cref{fig:blur}. While the imposed edge continuity is counter-productive with respect to the noise-free image in this example, we noticed that human image segmentations often rely on the principle of edge continuity.

\clearpage
{\small
\bibliographystyle{ieee_fullname}
\bibliography{egbib}
}